\pdfoutput=1
\documentclass[11pt]{article}
\usepackage[preprint]{acl}

\usepackage{times}
\usepackage{latexsym}

\usepackage[T1]{fontenc}
\usepackage{parskip}
\usepackage[utf8]{inputenc}

\usepackage{microtype}

\usepackage{inconsolata}
\usepackage{xcolor}

\usepackage{graphicx}

\title{Humanizing Machines: Rethinking LLM Anthropomorphism Through a Multi-Level Framework of Design}

\author{
    Yunze Xiao$^{*}$, Lynnette Hui Xian Ng\thanks{Yunze Xiao and Lynnette Hui Xian Ng contributed equally to this paper}, Jiarui Liu, Mona Diab\\ 
    Carnegie Mellon University \\
    \texttt{\{yunzex,huixiann,jiaruil5,mdiab\}@cs.cmu.edu} 
}

\begin{document}
\maketitle
\begin{abstract}
Large Language Models (LLMs) increasingly exhibit \textbf{anthropomorphism} characteristics -- human-like qualities portrayed across their outlook, language, behavior, and reasoning functions. Such characteristics enable more intuitive and engaging human-AI interactions. However, current research on anthropomorphism remains predominantly risk-focused, emphasizing over-trust and user deception while offering limited design guidance. We argue that anthropomorphism should instead be treated as a \emph{concept of design} that can be intentionally tuned to support user goals. Drawing from multiple disciplines, we propose that the anthropomorphism of an LLM-based artifact should reflect the interaction between artifact designers and interpreters. This interaction is facilitated by cues embedded in the artifact by the designers and the (cognitive) responses of the interpreters to the cues. 
Cues are categorized into four dimensions: \textit{perceptive, linguistic, behavioral}, and \textit{cognitive}. By analyzing the manifestation and effectiveness of each cue, we provide a unified taxonomy with actionable levers for practitioners. Consequently, we advocate for function-oriented evaluations of anthropomorphic design.
\end{abstract}

\section{Introduction}
Anthropomorphism is a purposeful design strategy that unlocks richer, more intuitive collaboration between humans and Artificial Intelligent (AI) systems. Developers equip large language models (LLMs) with relatable personalities  \cite{wang-etal-2024-incharacter,huang2024chatgptbenchmarkingllmspsychological}, emotional expressiveness  \cite{NEURIPS2024_b0049c3f}, and context-sensitive social reasoning  \cite{nighojkar2025givingaipersonalitiesleads,liu2025syntheticsocraticdebatesexamining}. These human-like cues allow users to converse with a system in familiar terms. This design principle builds user trust by reducing the cognitive effort required to interact with the system. Multi-modal extensions amplify this effect and create seamless and engaging interactions: speech synthesis models convey nuanced emotion
 \cite{zhou2022speechsynthesismixedemotions}, embodied agents navigate physical space
 \cite{xie2025embodiedraggeneralnonparametricembodied}, and vision-based models interpret social scenes
 \cite{mathur2025socialgenomegroundedsocial}. These human-like agents have shown tangible benefits to our society: providing realistic training environments for education  \cite{ma2024studentsexpertsnewai}, improving adherence and empathy for virtual healthcare consultations  \cite{Wen_2024}, strengthening therapeutic alliances in psychiatry \cite{wang-etal-2024-patient}, or making legal reasoning tools accessible to non‑experts  \cite{huang2023lawyerllamatechnicalreport}. In each scenario, carefully calibrated anthropomorphic cues bridge complex AI capabilities and human goals, enabling technology that feels / appears to be supportive, transparent, and responsive.

However, current work on anthropomorphism in LLM design is framed predominantly through a risk-centric lens. This lens emphasizes user misconceptions about model capabilities  \cite{tejeda2025what}, misplaced trust in dialog systems  \cite{zhou2025relai}, and the danger of over-reliance on emotions  \cite{Akbulut_Weidinger_Manzini_Gabriel_Rieser_2024}. This context has cultivated a cautious and often skeptical stance towards anthropomorphism within the community, often citing incidents where users disclosed personal financial information to chatbots they perceived as trustworthy humans  \cite{mireshghallah2024trustbotdiscoveringpersonal}, or cases where AI assistants reinforced harmful stereotypes through personality-driven responses  \cite{liu-etal-2024-evaluating-large}. While these concerns are legitimate, recent research on anthropomorphism has been largely shaped by its perceived harms and discouraged a deeper exploration of its functional or context-sensitive benefits \cite{olteanu2025aiautomatonsaisystems}. This dominant discourse leaves little room for feature-driven inquiry into when, how, and for whom anthropomorphic elements might enhance usability, trust calibration, or engagement in NLP applications.

\begin{figure*}[ht]
    \centering
    \includegraphics[width=1.1\linewidth]{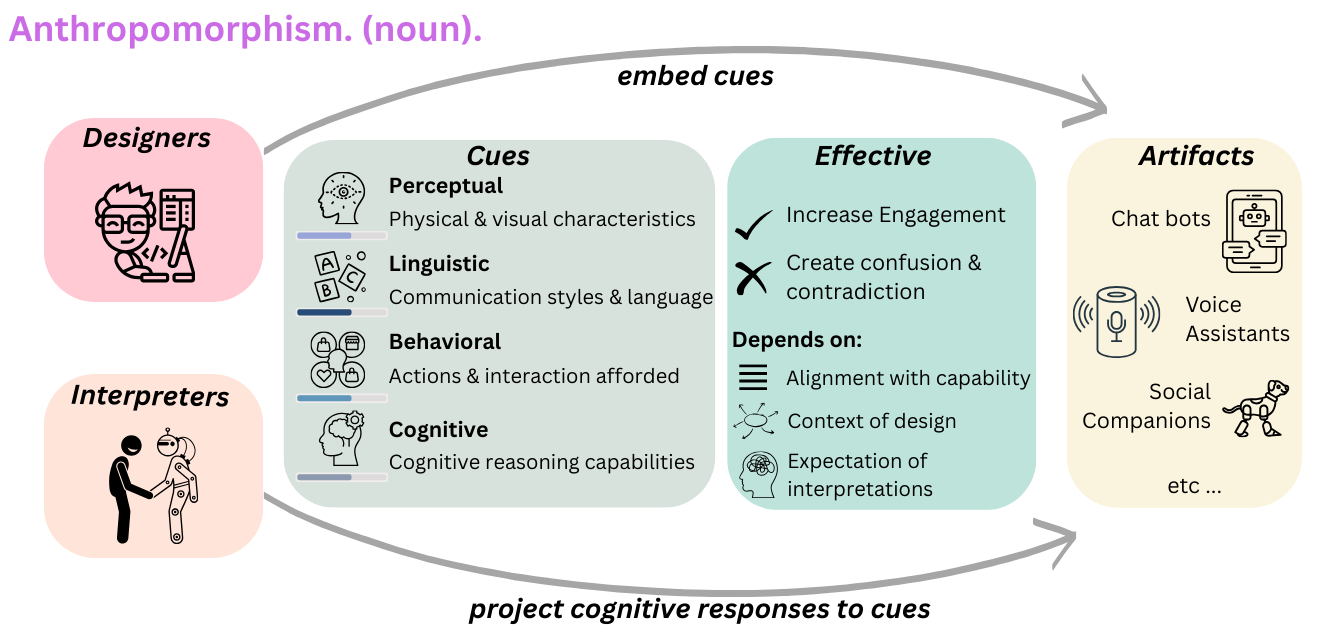}
    \caption{Illustrative definition of Anthropomorphism}
    \label{fig:illustration}
\end{figure*}

To address this gap, we propose a new definition of anthropomorphism which is grounded in how today's LLM-centered systems occupy the liminal space between utilitarian tools and social actors and moves beyond the defensive, risk-centric paradigm. Instead of treating anthropomorphism solely as a linguistic attribution of human-like characteristics to non-human entities  \cite{cheng2025humtdumtmeasuringcontrolling,cheng-etal-2024-anthroscore,DeVrio_2025}, we contend that anthropomorphism should be treated as a multidimensional reciprocal interaction process. Designers intentionally embed human-like cues into AI systems. Interpreters are users who, in turn, project their agency and mental states onto this system. 

This multimodal reconceptualization of anthropomorphism expands the research agenda, enabling scholars to systematically investigate the scenarios in which anthropomorphic design cues are beneficial. This conceptualization is critical because LLMs increasingly mediate experiences in sensitive and subjective domains \cite{ng2025aurasightgeneratingrealisticsocial}, in which the trust and engagement of interpreters profoundly shape the outcomes. The nuanced understanding of anthropomorphism is based on the interaction between user and model and focuses on the effectiveness of system design. This approach reduces mis-characterizations of system effects through over-reliance of cautionary narratives, and the deployment of human-like systems without adequate design foresight.
Our definition thus provides a coherent frame for auditing human-imitative AI across varied modalities, bringing a new perspective to harness anthropomorphism responsibly and effectively.

Our contribution is threefold: 
\begin{enumerate}
    \item We propose a new definition of anthropomorphism for today's NLP systems. LLMs are more than just tools; they are also social partners. The usage of LLMs depends on how designers build them and how interpreters respond and interact;
    \item We advocate for a shift from the prevailing risk-centric evaluations of anthropomorphism towards an effectiveness-focused approach. We analyze existing studies for actionable insights into anthropomorphic design decisions;
    \item We present a feature-driven design framework based on the reactions of interpreters to human-like features built into LLM systems by designers.
    \end{enumerate}

\section{What is Anthropomorphism?}

Anthropomorphism has long been a fundamental phenomenon in the study of human–machine interaction. As early as 1950, Turing’s Imitation Game  \cite{turing1950computing} framed the ability of machines to mimic human behavior as a measure of intelligence. In 1966, Weizenbaum introduced the "ELIZA effect", in which simple linguistic cues can elicit deep emotional responses from interpreters  \cite{weizenbaum1966eliza}. In 1970, the Japanese roboticist Mori introduced the concept of the Uncanny Valley and showed that increasing the realism of robots can cause unease to human interpreters when the artificial agents do not completely resemble humans  \cite{6213238}. 

Adjacent fields to NLP such as Human-Computer Interaction and Information Sciences have developed rich accounts of anthropomorphism as a socio-technical phenomenon shaped by design intentions, interpreter expectations, and context \cite{frazer2022experimental,damholdt2023scoping}. These well-established traditions provide guidance towards a context-aware understanding of anthropomorphism, one that can inform both the design and the evaluation of human-like language technologies.

Contemporary NLP debates often treat anthropomorphism either as a narrow linguistic phenomenon or as a hazard requiring mitigation, frequently foregrounding the risks of over-trust, deception, or disinformation while sidelining potential design benefits. Much of the literature has emphasized preventing misleading cues or curbing anthropomorphic projections  \cite{doi:10.1073/pnas.2415898122}

To restore this missing depth and anchor our multi-level framework in conceptual rigor, we first synthesize peer-reviewed definitions of anthropomorphism drawn from Robotics, Human-AI Interaction (HAI), and Natural Language Processing (NLP). This synthesis is not merely classificatory; it underpins our argument that anthropomorphism should be understood as a context-sensitive interaction between designers, systems, and interpreters rather than as a fixed set of traits. The NLP field currently lacks a comparable design-oriented taxonomy for anthropomorphism. Current frames of anthropomorphism are heavily risk-oriented, elaborating on trust and safety issues that may arise with heavily anthropomorphic systems. Our literature synthesis in Section~\ref{sec:app_definitions} serves as an implicit comparative analysis, demonstrating how our framework integrates and organizes fragmented perspectives from adjacent fields.

To carry out this synthesis, we adopted a diachronic review strategy, sorting definitions into three eras that align with the major technological and conceptual shifts within each domain. This historical approach traces converging themes and exposes disciplinary blind spots across Robotics, HAI, and Information Science. Detailed periodization, selection criteria, and source lists are provided in the appendix \ref{sec:app_definitions}.

To better understand anthropomorphism in the context of LLMs, we first map the ecosystem in which it arises: the human‑AI interaction space comprising four core components that operate in tandem:
\begin{enumerate}
    \item \textbf{Artifacts} are the AI systems themselves, the medium for interactions. Examples: LLM chatbots, voice assistants, or social robots;
    \item \textbf{Cues} are perceptual, linguistic, behavioral, or cognitive signals built into artifacts to trigger human-like readings. Cues can be deliberately or inadvertently built into the system. Examples: a humanoid silhouette, a sympathetic tone, a first‑person pronoun. \autoref{ref:design} presents the design principles of the cues;
    \item \textbf{Designers} create the artifacts. They embed cues that influence the perception and interaction of the artifact;
    \item \textbf{Interpreters} are the human users of the artifacts. Interpreters, driven by their mental states (intentions, emotions, and agency), project their cognitive response to the cues onto the artifact.
\end{enumerate}

Thus, anthropomorphism in the context of LLM is defined as follows:
\begin{quote}
    Anthropomorphism is a reciprocal phenomenon in which \textbf{ designers} embed human-like \textbf{cues} into \textbf{artifacts}; and \textbf{interpreters} project their cognitive response to the cues onto the artifacts. 
\end{quote}

Designers may purposefully embed anthropomorphism, whereas interpreters usually are driven by their purposes and intentions. These \textbf{purposes} are the underlying intentions and contextual needs that drive the design and interpretation of the LLM output. Purposes connect between the four core elements of anthropomorphism and shape the system's implementation and perception.
    \begin{enumerate}
        \item From the \textbf{designer's perspective}, purposes include: enhancing usability, fostering interpreter trust, encouraging participation,  or simulating companionship.
        \item From the \textbf{interpreter’s perspective}, purposes arise from social, emotional, or functional needs. Examples: the desire for empathy, efficiency, or human-like interaction.
    \end{enumerate}

Our definition encompasses a range of \textbf{cues} that designers can supply and the corresponding \textbf{projections} that interpreters can generate. An artifact does not necessarily incorporate all the cues. Neither will every interpreter attribute the same mental states. 
Rather, the \textit{mode} and \textit{function} of anthropomorphism at work is given by the \textbf{unique mix of cues provided and projections obtained} from the AI system.

\section{Manifestation of Anthropomorphism}
\label{ref:design}
%The manifestation of anthropomorphism lies in the design principles of \textit{Cues}. The cues are categorized along four axes, which are identified through extracting similarities from the literature definitions, then refined through extensive team discussions grounded in empirical examples, theoretical frames and speculative design scenarios. These cues align themselves along the established Theory of Mind (ToM) framework  \cite{wellman1990child},which describes the human ability to attribute mental states to others, a key attribute of anthropomorphism (\autoref{tab:tom_dimensions}).
The manifestation of anthropomorphism lies in the design principles of \textit{Cues}. These cues are categorized along four axes, identified through a synthesizing of converging themes in existing definitions from the literature, and then refined through extensive team discussions. Each axis represents a different type of anthropomorphic cue and can vary in intensity, indicating how strongly human-like traits are expressed in the system. These axes align with the established Theory of Mind (ToM) framework  \cite{wellman1990child}, which describes the human ability to attribute mental states to others, a key attribute of anthropomorphism (\autoref{tab:tom_dimensions}).

\autoref{fig:illustration} illustrates the design principles of the four cues: 1) Perceptual (\ref{sec:perceptual}), captures physical or visual elements that convey human-likeness; 2) Language (\ref{sec:lang}), encompasses communication styles to signal the degree of humanness; 3) Behavioral (\ref{sec:beh}), describes actions, responses, and interaction patterns; and 4) Cognitive (\ref{sec:cog}), refers to the cognitive reasoning capabilities attributed to artifacts. Importantly,each dimension functions on a continuum from low to high anthropomorphism. Low-level cues have minimal human-like characteristics that evoke basic social responses without requiring complex implementation. High-level cues approximate human traits more closely, generating sophisticated cognitive and emotional responses, naturally necessitating more advanced design techniques. We treat the aggregate intensity of perceptual, linguistic, behavioral, and cognitive cues as a calibrated parameter $\alpha$ that designers can dial up or down to match the artifact's system competence.

\begin{table}[ht]
\centering
\caption{Alignment Between Our Anthropomorphic Dimensions and Wellman's Theory of Mind (ToM) Components}
\begin{tabular}{ll}
\hline
\textbf{Design Principles} & \textbf{ToM Dimension} \\
\hline
Perceptual & Perceiving \\
Linguistic & Feeling + Desiring \\
Behavioral & Choosing \\
Cognitive & Thinking \\
\hline
\end{tabular}
\label{tab:tom_dimensions}

\end{table}
\subsection{Perceptual Cues}
\label{sec:perceptual}
The Perceptual dimension refers to the physical or visual features of an artifact that conveys a sense of human-likeness to interpreters. These features contribute to the interpreters' first impression of the artifact. Perceptual anthropomorphism can be understood through a spectrum of low-level to high-level cues. Low-level cues are generic representations (for example, a standardized avatar) and abstract symbolism (for example, two dots to represent eyes). High-level cues are personalized representations (e.g., an avatar face modeled after a known individual) that have extremely realistic detail (e.g., anatomically accurate facial musculature). 

The specificity and intensity of the perceptual cues can be mapped onto a continuum that spans from minimal abstraction to high realism. Some systems exhibit high intensity but low specificity, therefore appearing very human-like without resemblance to real-life figures. An example is Geminoid-F created by Hiroshi Ishiguro  \cite{5953147}. Other systems are highly specific, but are mildly human-like, such as stylized avatars.

The perceptual continuum not only affords the aesthetic features of an artifact, but also governs the cognitive mechanism activated in interpreters. As an artifact moves from abstract to realistic representations, the degree to which interpreters project their mental states increases  \cite{paivio1978mental}. This progression reflects a shift from an object-like characterization to an agent-like engagement, to a system perceived as a social or emotional presence.

Perceptual anthropomorphism has significant implications for system deployment. Human-like visual cues inherently shape interpreter expectations and engagement. Even masked features such as built-in sensors or subtle gestures can elicit basic social responses to turn or share attention  \cite{urakami2022nonverbal}. As the realism of these cues increases, interpreters attribute more complex qualities to the system and foster HAI interactions grounded in trust, empathy, or emotional dependence  \cite{tu2023effects}. However, when the human-like appearance of the artifact exceeds its communicative or cognitive capabilities, interpreters often experience cognitive dissonance  \cite{yu2023cognitive}. This mismatch between expectations and reality leads to discomfort or rejection of the system effect \cite{6213238}. Perceptual cues establish interpretive frames and hence must be aligned with system competence for realism. This consistent principle has been reiterated in multiple domains, ranging from robotics to NLP  \cite{6213238,kang2025mitigating}. Designers must therefore calibrate the level of perceptual realism to preserve coherence in interpreter projections, thereby maintaining effective HAI interaction. 

\subsection{Linguistic Cues}
\label{sec:lang}
The linguistic dimension refers to the use of language to shape interpreter perceptions. At its core, this dimension captures how specific linguistic choices signal humanness. These linguistic cues encompass elements of vocabulary, syntax, and tone that manifest themselves in choices of pronoun usage, degree of formality, or emotional expression. Such cues are commonly analyzed in computational social sciences to provide insight into the writer's psychological state and the receiver's reactive interpretations  \cite{pennycook2020falls}. For example, the use of reassurance and personal pronouns increases the trust of the interpreter in turn-based conversations  \cite{jaidka2024takes}. Importantly, these linguistic markers serve as surface realizations, such as choices of pronouns, hedges, and affective words, that may imply agency or mental states without requiring the model to actually possess such capabilities.

When artifacts adopt human-like language patterns, they activate the social-cognitive schema in interpreters  \cite{weizenbaum1966eliza}. Low-level linguistic cues refer to superficial markers of social interaction, such as the use of personal pronouns ("I"), hedges ("maybe“), or politeness strategies ("please"). High-level linguistic cues involve more complex discourse behaviors, which from a surface level, implies some level of cognitive abilities.  Such cues are used to justify actions, make inferences, or manage conversational dynamics. Artifacts that engage in seamless conversational turn-taking typically exhibit high-level cues, whereas those designed primarily for information delivery tend to rely on low-level ones. Importantly, these cues do not necessarily indicate true cognitive capabilities (cf. Section \ref{sec:cog}); rather, they serve as surface-level representations of such agency without requiring the model to substantiate its implied mental states. 

Several recent studies have proposed valuable systematic methods to quantify the degree of linguistic anthropomorphism in LLMs  \cite{cheng2025humtdumtmeasuringcontrolling,cheng-etal-2024-anthroscore,DeVrio_2025}. These works introduce metrics such as HUMT and AnthroScore that use linguistic features and conversational alignment to assess how human-like an artifact's language output appears. 

However, the impact of linguistic cues on interpreters is not uniform. Interpreter responses are moderated by individual priors \cite{abercrombie-etal-2023-mirages,Basoah2025NotLU}, system environments \cite{sah2015effects}, and cultural background factors  \cite{alkhamissi-etal-2024-investigating,eyssel2015cultural}. Language choices thus not only modulate the tone of the interaction, but also frame the perceived role and competence of the artifact. Therefore, designers must carefully calibrate linguistic anthropomorphism to align with both the artifact's intended function and the target audience's expectations.  This calibration involves considering cultural differences and adapting linguistic strategies based on the artifact's purpose. Designers must balance human-like language with clear signals of the artifact's nonhuman nature while regularly evaluating how such choices affect user trust. The goal is not maximum human-likeness. Instead, designers should create language patterns that establish appropriate mental models. 

\subsection{Behavioral Cues}
\label{sec:beh}
The behavioral dimension refers to the actions and interaction patterns that the artifact affords. Behavior dynamically connects (1) the designer's embodiment of the artifact with (2) the interpreter's expectations of the artifact's behavior and their intentionality of use. Artifacts demonstrate behavior through contingent responses (that is, answering a question) \cite{yue2025surveylargelanguagemodel}, proactive pursuit of goals (that is, formulating a travel itinerary) \cite{10.5555/3692070.3694316}, or adaptive interaction (that is, personalized results)  \cite{chen2024personapersonalizationsurveyroleplaying}. 

Designers embed behavioral cues to a varied degree. Embodied agents that have a physical or simulated presence (e.g., social agents, avatars) have a high level of behavioral cues. These artifacts signal behavior through physical gestures, spatial coordination, and environmental manipulation. Nonembodied agents that operate entirely within digital environments (e.g., web-based chatbots, coding assistants) have low level of behavioral cues and operationalize behavior through online actions that do not have a physical referent like API calls and autonomous task planning. 

Behavioral cues activate cognitive mechanisms in interpreters that attribute mental states to artifacts upon the display of certain behavioral signatures  \cite{urquiza2015mind,marchesi2022mental}. Interpreters vary the number of behavioral cues they desire in each setting of the environment. Environments such as AI companions are expected to have more behavioral cues for goal-directed actions and emotional adaptability. Task-specific environments such as code autocompletion can have lower levels of behavioral cues (e.g., providing functionally contingent responses).

Behavioral affordances guide interpreters towards projecting human-like qualities onto the artifact. More adaptive, timely, and seemingly intentional behaviors result in a higher likelihood of interpreters casting human-like traits like emotion or social presence onto the artifact. Designers can leverage on this behavioral dimension and its interaction with the artifact's capabilities and environmental context to more precisely shape the interpreter's cognitive response toward the artifact and calibrate expectations around trust, competence, and companionship. 

\subsection{Cognitive Cues (reasoning level signal) }
\label{sec:cog}
The cognitive dimension refers to the cognitive reasoning capabilities of the artifact. This includes: (1) the abilities embedded by the designer, such as the ability to reflect, plan, learn, or make inferences; and (2) the abilities perceived by the interpreter, such as self-correction, expressing uncertainty, or adapting responses  \cite{guo2025deepseek}.

Cognitive cues operationalized through system behaviors that suggest internal system deliberation or state modeling\cite{xu2025guessithinkingbenchmark}, even when no such process exists. LLMs reflect the cognitive dimension by alluding to thinking behavior with their outputs that simulate reasoning, reflection, or uncertainty. Such outputs serve as cognitive signals that prompt the interpreter to assign intelligence or thoughtfulness to the system. 

The cognitive cues of an artifact can be further distinguished through the complexity, consistency, and transparency of the cues. High-level cognitive cues emerge when artifacts display complex, dynamic behaviors that closely mirror human cognitive reasoning. High-level cues simulate the appearance that the artifact can monitor and adjust its thinking, leading to interpretations of an intelligent or self-aware system. Such cues include: the display of complex emotions like empathy  \cite{huang2024emotionallynumbempatheticevaluating} , sophisticated logical reasoning like math problems  \cite{tsoukalas2024putnambenchevaluatingneuraltheoremprovers} or the carrying out of elaborate conversational tasks like negotiations  \cite{jaidka2024takes}. Low-level cognitive cues are behaviors that provide a small hint of mental activity. Such behaviors suggest minimal embedding of cognitive cues and are less likely to trigger strong mental state attribution  \cite{coricelli2005two}. This includes token expressions of reasoning or uncertainty or stating a variant of an input prompt. 

The implications of cognitive anthropomorphism are highly context-dependent. In the contexts of education and mental health, common artifacts are conversational chatbots. In these artifacts, high-level cognitive cues such as empathy expression, reflective revision\cite{yang2025ahamoments}, and logical reasoning, can enhance the interpreted competence of the artifacts. When interpreters view the artifacts as more thoughtful and emotionally aware, there is increased engagement and trust  \cite{gillath2021attachment}. However, in the context of web search and other task-specific applications, low-level cues that promote clarity and efficiency will suffice. Designers should thus calibrate the degree of cognitive cues to the context of artifact use in order to increase usability and user satisfaction.

\section{When Are Anthropomorphic Cues Effective?}
The effectiveness of anthropomorphic cues depends critically on three alignment factors: (1) \textbf{capability-expectation} alignment, which ensures cues don't promise more/less than the user expects, (2) \textbf{context-purpose} alignment, which matches cue intensity to the interaction's stakes and requirements, and (3) \textbf{cultural-norm} alignment, which ensures cues respect diverse interaction expectations. When these alignments break down, the same cues that enhance engagement can produce documented harms including over-trust, emotional manipulation, and dangerous over-reliance. The following analysis examines both positive and negative outcomes for each cue dimension, with particular attention to design strategies that maintain beneficial effects while mitigating predictable risks.

\subsection{Perceptual Cues}
Perceptual cues such as the visual embodiment or the name of the artifact, are powerful levers as first impressions to shape interpreter expectations for LLM-based systems. Effective perceptual cues, such as realistic avatars, can improve trust and usability, especially when those cues align with the artifact's competence  \citep{chattopadhyay2017perceptual,kulms2023anthropomorphic,moore2024uncanny}. In LLM interfaces, these cues can be friendly greetings or typing animations  \citep{goyal2024designing,kulms2016anthropomorphic}. 

Perceptual cues backfire when they suggest cognitive depth or social intelligence beyond what the artifact can reliably deliver. For example, overly realistic avatars can lead interpreters to overestimate the model's capabilities, which can result in disappointment or trust erosion when the artifact does not meet expectations  \citep{crolic2020chatbot,chattopadhyay2017perceptual}. LLM-based artifacts that mimic human conversational patterns are especially vulnerable to  \textit{anthropomorphic projection bias}, a phenomenon where interpreters punish them severely for errors or shallow reasoning  \citep{jiang2022trust}. This mismatch occurs when initial perceptual cues activate intuitive trust, but subsequent interaction exposes a lack of deeper understanding of the problem  \citep{eyssel2021trust,nass1997anthropomorphism}.

\subsection{Linguistic Cues}

Well-chosen linguistic devices can strengthen user trust and rapport. Empirical work shows that first-person pronouns ("I") and emotive language increase perceived credibility and lower perceived risk in LLM outputs  \cite{velner2021speaking,cohn2024believinganthropomorphismexaminingrole,ibrahim2025multiturnevaluationanthropomorphicbehaviours}. However, the same cues backfire when the system under-delivers: anthropomorphism inflates expectations, so a single failure can yield sharper anger and lower satisfaction  \cite{carter2023meaningful,crolic2022blame}. In role-play settings, stylistic choices can also amplify social biases  \cite{liu-etal-2024-evaluating-large}.

To decide how much human-like language to embed, designers should quantify cue intensity with metrics such as HumT and AnthroScore  \cite{cheng2025humtdumtmeasuringcontrolling,cheng-etal-2024-anthroscore}. These measurements, coupled with iterative cross-cultural user tests, help align linguistic style with task goals and audience expectations, preventing over- or under-anthropomorphism. Designers should also conduct regular cross-cultural user tests to verify that cue wording is perceived as natural and respectful. For example, the hedge "maybe" reads as a polite mitigation to Americans but as evasive to Koreans  \cite{Duffau2024ExpectingPP,yu2011indirectness}.

%Linguistic cues become ineffective under misaligned trust calibration.  \citeauthor{carter2023meaningful} demonstrates that anthropomorphism can elevate subjective trust but can lead to dangerous miscalibrations should the artifact fail to perform as predicted. For example, as  \citeauthor{crolic2022blame} have shown that providing linguistic cues indeed raised pre-interaction trust, but after a single service failure those same cues triggered significantly more anger, lower satisfaction, and harsher firm evaluations. Linguistic cues, especially in role-play scenarios, might also elicit a strong correlation with social signals that perpetuate harmful biases and therefore must be properly calibrated  \cite{liu-etal-2024-evaluating-large,gupta2024personabias}.

%To better utilize linguistic cues, designers can use available linguistic metrics like HumT and AnthroScore \cite{cheng2025humtdumtmeasuringcontrolling,cheng-etal-2024-anthroscore} to evaluate the extent of linguistic cue usage required for their envisioned context. Designers should continually perform user testing to ensure that interpreters are comfortable with the language usage and style of the cues output from the artifact. These user tests should also be performed across diverse cultures to account for language-based cultural nuances. 

\subsection{Behavioral Cues}
Behavioral cues such as following social norms in responses and adjusting responses to the situation serve as a mirage for artifacts to match the interpreter's contextual expectations. For example, conversational bots that follow the conversation flow are generally rated as more engaging and trustworthy  \cite{yang2024}. In assistive contexts such as code generation, shared autonomy behavior enhances both task performance and interpreter satisfaction  \cite{10.1145/3586030}. In such contexts, norm-adaptive behaviors (that is, adapting turn-taking latency and politeness markers to local sociocultural conventions) that modulate cultural and turn-taking conventions improve the acceptance of artifacts  \cite{eyssel2015cultural}.

Behavioral cues are counterproductive when they imply unjustified autonomy or enforce rigid and biased norms  \cite{schramowski2021large,parsons2023ethics}. Overly proactive chatbots that make unsolicited decisions, interrupt conversations, or provide extremely long responses are often perceived as intrusive  \cite{10.1145/3613905.3650823,reicherts2021interrupt}. These behaviors undermine the autonomy of interpreters, leading to discomfort and disengagement of interaction with the artifact. Normative biases embedded in artifact behavior can further exacerbate these boundary violations  \cite{parsons2023ethics}. 
These issues have to be taken into account when designing artifacts for open-world deployments where social dynamics cannot be easily codified  \cite{pinch1984social}.

\subsection{Cognitive Cues}
Cognitive cues are effective when they simulate mental processes that align with interpreter expectations about reasoning, understanding, and adaptation. Cognitive cues as expression of empathy and autonomous error correction are particularly effective in relationship-oriented settings. Examples of such settings are emotional support chatbots or social companionship bots  \cite{Park2022,Lee2024,DeGennaro2020,Ehrlich2023,Luo2025}. Different types of cognitive cues, such as basic error acknowledgments and providing uncertainty statements, can meaningfully support productivity-focused tasks such as core review. The appropriate use of cognitive cues tempers overexpectations on the artifact's outputs and maintains stable trust during repetitive and high-stakes scenarios  \cite{Kim2024}.

In high-stakes domains affecting human lives, anthropomorphism serves a critical interpretability function beyond mere trust maintenance. When LLMs make decisions in medical ethics or legal contexts, human-like reasoning patterns enhance transparency and accountability. For example, a medical AI explaining treatment recommendations using familiar ethical frameworks (beneficence, autonomy) makes its decision process more interpretable to healthcare providers. Similarly, legal AI systems that articulate reasoning through precedent and principle mirror human judicial thinking, enabling meaningful oversight  \cite{Kim2024}.

Cognitive cues backfire when the artifact displays greater depth than necessary. For example, overly emotional displays in healthcare chatbots can reduce authenticity and mislead vulnerable populations (e.g., children, older adults) into over-trusting the system  \cite{Seitz2024}. On the other hand, shallow, seemingly empathetic phrases without adaptive reasoning are quickly judged insincere  \cite{Lee2024,Liu2024}. Likewise, artifacts that issue unsolicited recommendations or do not recover from mistakes erode the confidence of the interpreter, especially in unpredictable environments  \cite{Stiber2025}.

\section{Recommendations}
Based on our multidimensional framework and effectiveness analysis, we offer the following recommendations for practitioners designing LLM systems with anthropomorphic elements:
\begin{table*}[h]
\centering
\caption{Anthropomorphic Cues: Context-Dependent Applications}
\begin{tabular}{lll }
\hline
\textbf{Cue Type} & \textbf{Beneficial Use} & \textbf{Minimize When} \\
\hline
Cognitive & Mental health (empathy, reasoning) & Search engines (trust misleading) \\
Linguistic & Education (conversational) & Finance (transaction seriousness) \\
Behavioral & Social AI (rapport building) & Legal bots (false authority) \\
Perceptual & Children's learning (friendly avatar) & Government (official identity) \\
\hline
\end{tabular}
\label{tab:example}
\end{table*}
\paragraph{Align cues with artifact capabilities:} 
Effective anthropomorphic design requires carefully calibrating embedded cues to match the actual capabilities of the artifact. Perceptual cues should be proportionate to the system's reliability and the criticality of the task that the artifact supports  \citep{kulms2023anthropomorphic}. Overly human-like features can create false expectations, leading to disappointment when performance falls short. Designers must also adapt perceptual cues such as color, gestures, and politeness to cultural norms to avoid misinterpretation  \citep{eyssel2015cultural}. Beyond perception, behavioral cues should be aligned with the demands of the interpreter-artifact interaction and the interpreter's preferences. This behavioral calibration should be dynamic and adjusted to environmental and contextual changes. Artifact design should be informed by the input of various stakeholder communities to avoid normative misalignment and cultural insensitivity  \citep{olteanu2025aiautomatonsaisystems}. This should be taken into account especially in the design of linguistic cues, because interpreters expect different levels of linguistic cues in different scenarios. Finally, cognitive cues must reflect what the system can genuinely deliver. Designers should avoid simulating complex reasoning if the algorithms underlying the artifacts lack such capabilities, as such this can result in miscalibrated trust. The different cues should be paired together with transparent user interfaces to signal the artifact's functional boundaries and affordances. 

\paragraph{Participatory implementation techniques:} Anthropomorphic features should be implemented on a sliding scale that supports adaptive participatory anthropomorphism, where the system anticipates the intensity of the preferred signal while preserving the user's override. This means creating artifacts that learn and adjust based on interpreter preferences over time. The more control interpreters they have across the four cue dimensions, the more precisely the systems can align with individual expectations and needs. Designers should embed adjustable parameters to empower interpreters and maintain transparency. Explicit markers of reasoning processes and feedback mechanisms can further support interpreter trust. By dynamically updating the anthropomorphic profile in response to interaction data, artifacts become more personalized and more aligned with their functional characteristics. Crucially, this approach calls for ongoing personalization of LLM, where systems continuously adapt to the evolving communication norms, emotions, and moral expectations of users  \cite{wang2024aipersonalifelongpersonalization}.

\paragraph{Context-sensitive implementation:} Anthropomorphic design should not be treated as a one-size-fits-all solution; rather, it must be calibrated to the specific context in which an artifact is deployed. Designers should visualize and monitor the evolution of interpreter-artifact dynamics across repeated interactions, potentially using metrics such as trust calibration, emotional attribution, and perceived autonomy. Anthropomorphic intensity might need to start high to facilitate initial engagement, but should ideally be adaptive. Design artifacts with sensitivity to cultural variation, as signals that build trust in one culture may cause discomfort in another. The cultural context should be treated as a dynamic design variable. The NLP community should engage more deeply in cultural adaptation, drawing inspiration from work such as  \citeauthor{shiomi2024bowing}.  Examples are shown in \autoref{tab:example}.

\paragraph{Future-oriented Evaluations:} Anthropomorphic interfaces should be viewed as evolving entities that advance alongside LLMs and shifts in public expectations. Designers therefore need reliable metrics for each anthropomorphic dimension. Currently, indices such as HUMT quantify linguistic cues, but measures for behavioral and cognitive cues are still lacking for NLP researchers and form a crucial research avenue. Once a complete set of metrics exists, teams should correlate them with task outcomes (i.e. success rate, error frequency) to determine whether increased human-likeness improves performance and to adjust cue intensity for optimal benefit and risk. Creating and validating these metrics requires collaboration between HCI, social psychology, and cultural anthropology so that the concept of humanness respects diverse norms. Through this interdisciplinary and data-driven process, designers can ensure that anthropomorphic features remain socially clear and ethically appropriate as both technology and culture continue to evolve.

\section{Conclusion}
Our paper approaches the concept of anthropomorphism as a calibrated parameter resulting from the design and interpretation of an artifact. We categorize this phenomenon as the embedding and projection of responses across perceptual, linguistic, behavioral, and cognitive cues. Drawing from research of adjacent fields, we show how calibrated anthropomorphic features could increase engagement when aligned with the artifact's capability, design context, and interpretation expectations. By applying this concept to LLMs, we show how balancing technical design considerations and user expectations should allow LLMs to serve both as tools and as social partners.

While our framework analytically separates these four cues, real deployments often feature overlapping and mutually reinforcing interactions. Future research should theorize and model these interdependencies,  for instance, how perceptual realism enhances the credibility of linguistic output, or how cognitive signals of reasoning influence perceptions of behavioral adaptability. Rather than treating cues in isolation, scholars should formalize their joint dynamics to capture cross-cue reinforcement, compensation, or interference, providing deeper insight into the systemic nature of anthropomorphic design. 

\section*{Limitations and Ethical Considerations}
Our work presents a conceptual framework to understand anthropomorphism in LLM-based artifacts through four dimensions: perceptual, linguistic, behavioral and cognitive cues. These four dimensions are embedded in an artifact by a designer and responded to by interpreters. Although we believe this taxonomy offers practical design guidance to anthropomorphic artifacts, it is important to recognize several limitations and ethical considerations.

\textbf{Limitations.} This study is primarily theoretical and synthesizes insights from the previous literature in NLP, HCI, and robotics. We did not conduct empirical evaluations, user studies, or automated cue quantification at scale. As such, our claims are not validated through direct user interaction or system testing. Although we propose a multilevel cue framework, real-world deployments often feature overlapping or entangled modalities (e.g., linguistic and cognitive cues co-occurring in emotionally expressive dialogue). Our framework idealizes these dimensions for analytical clarity, which may limit its robustness when applied to noisy, mixed-modality systems. Moreover, the proposed framework assumes that designers have control over the degree and type of anthropomorphic cues presented, which may not hold in black-box or commercial LLM deployments. 

\textbf{Ethical Considerations and Potential Risks.} Anthropomorphic design, if misaligned with actual system capabilities, can lead to mis-calibrated trust, user over-reliance, or affective misinterpretation. This is particularly of concern in emotionally sensitive domains such as healthcare, education, or companionship. Highly realistic cues can unintentionally signal cognitive or emotional competence that the artifact does not possess, raising risks of deception or exploitation of vulnerable user populations (e.g., children, elders). These risks are amplified when cues (e.g., cognitive cues of empathy expressions or apologies) are simulated without functional grounding, potentially undermining user autonomy and transparency.

Furthermore, anthropomorphic systems can entrench normative or cultural biases if behavioral and linguistic cues are not localized or participatory in design. This marginalizes underrepresented cultures or reinforces dominant interaction norms. Although our framework advocates for cross-cultural and context-sensitive cue calibration, more empirical research is needed to verify the effectiveness of such strategies in global deployments\cite{dai2025embracingcontradictiontheoreticalinconsistency}.

We also recognize the potential for dual use of our taxonomy. Our taxonomy could also be used to inform more persuasive or emotionally manipulative systems, especially in commercial, surveillance, or political contexts. We encourage future work to develop mitigation strategies, such as interpretability indicators, constrained anthropomorphic profiles, or gated release mechanisms, to help monitor and control anthropomorphic behavior. We also stress the importance of interdisciplinary collaboration with ethicists, domain experts, and affected communities during system development.

By articulating both the functional benefits and the possible harms of anthropomorphism in LLMs, our goal is to support transparent, socially aligned, and user-aware design practices. We strongly encourage future research to empirically validate and refine this framework, particularly through participatory co-design and cross-cultural evaluation.

\bibliography{custom,anthology_0}

\begin{thebibliography}{119}
\providecommand{\natexlab}[1]{#1}

\bibitem[{Abercrombie et~al.(2023)Abercrombie, Cercas~Curry, Dinkar, Rieser, and Talat}]{abercrombie-etal-2023-mirages}
Gavin Abercrombie, Amanda Cercas~Curry, Tanvi Dinkar, Verena Rieser, and Zeerak Talat. 2023.
\newblock \href {https://doi.org/10.18653/v1/2023.emnlp-main.290} {Mirages. on anthropomorphism in dialogue systems}.
\newblock In \emph{Proceedings of the 2023 Conference on Empirical Methods in Natural Language Processing}, pages 4776--4790, Singapore. Association for Computational Linguistics.

\bibitem[{Akbulut et~al.(2024)Akbulut, Weidinger, Manzini, Gabriel, and Rieser}]{Akbulut_Weidinger_Manzini_Gabriel_Rieser_2024}
Canfer Akbulut, Laura Weidinger, Arianna Manzini, Iason Gabriel, and Verena Rieser. 2024.
\newblock \href {https://doi.org/10.1609/aies.v7i1.31613} {All too human? mapping and mitigating the risk from anthropomorphic ai}.
\newblock \emph{Proceedings of the AAAI/ACM Conference on AI, Ethics, and Society}, 7(1):13--26.

\bibitem[{AlKhamissi et~al.(2024)AlKhamissi, ElNokrashy, Alkhamissi, and Diab}]{alkhamissi-etal-2024-investigating}
Badr AlKhamissi, Muhammad ElNokrashy, Mai Alkhamissi, and Mona Diab. 2024.
\newblock \href {https://doi.org/10.18653/v1/2024.acl-long.671} {Investigating cultural alignment of large language models}.
\newblock In \emph{Proceedings of the 62nd Annual Meeting of the Association for Computational Linguistics (Volume 1: Long Papers)}, pages 12404--12422, Bangkok, Thailand. Association for Computational Linguistics.

\bibitem[{Barke et~al.(2023)Barke, James, and Polikarpova}]{10.1145/3586030}
Shraddha Barke, Michael~B. James, and Nadia Polikarpova. 2023.
\newblock \href {https://doi.org/10.1145/3586030} {Grounded copilot: How programmers interact with code-generating models}.
\newblock \emph{Proc. ACM Program. Lang.}, 7(OOPSLA1).

\bibitem[{Basoah et~al.(2025)Basoah, Chechelnitsky, Long, Reinecke, Zerva, Zhou, D'iaz, and Sap}]{Basoah2025NotLU}
Jeffrey Basoah, Daniel Chechelnitsky, Tao Long, Katharina Reinecke, Chrysoula Zerva, Kaitlyn Zhou, Mark D'iaz, and Maarten Sap. 2025.
\newblock \href {https://api.semanticscholar.org/CorpusID:278481098} {Not like us, hunty: Measuring perceptions and behavioral effects of minoritized anthropomorphic cues in llms}.

\bibitem[{Becker-Asano and Ishiguro(2011)}]{5953147}
Christian Becker-Asano and Hiroshi Ishiguro. 2011.
\newblock \href {https://doi.org/10.1109/WACI.2011.5953147} {Evaluating facial displays of emotion for the android robot geminoid f}.
\newblock In \emph{2011 IEEE Workshop on Affective Computational Intelligence (WACI)}, pages 1--8.

\bibitem[{Blut et~al.(2021)Blut, Wang, W{\"u}nderlich, and Brock}]{Blut2021UnderstandingAI}
Markus Blut, Cheng Wang, Nancy~Viola W{\"u}nderlich, and Christian Brock. 2021.
\newblock \href {https://api.semanticscholar.org/CorpusID:230796429} {Understanding anthropomorphism in service provision: a meta-analysis of physical robots, chatbots, and other ai}.
\newblock \emph{Journal of the Academy of Marketing Science}, 49:632 -- 658.

\bibitem[{Boyer(1996)}]{boyer1996what}
Pascal Boyer. 1996.
\newblock \href {https://doi.org/10.2307/3034634} {What makes anthropomorphism natural: Intuitive ontology and cultural representations}.
\newblock \emph{Journal of the Royal Anthropological Institute}, 2(1):83--97.

\bibitem[{Burghardt(1991)}]{burghardt1991cognitive}
Gordon~M. Burghardt. 1991.
\newblock Cognitive ethology and critical anthropomorphism: A snake with two heads and hognose snakes that play dead.
\newblock In Carolyn~A. Ristau, editor, \emph{Cognitive Ethology: The Minds of Other Animals}, pages 53--90. Lawrence Erlbaum Associates, Hillsdale, NJ.

\bibitem[{Caporael(1986)}]{caporael1986anthropomorphism}
Linnda~R. Caporael. 1986.
\newblock \href {https://doi.org/10.1016/0747-5632(86)90004-X} {Anthropomorphism and mechanomorphism: Two faces of the human machine}.
\newblock \emph{Computers in Human Behavior}, 2(3):215--234.

\bibitem[{Carter et~al.(2023)Carter, Loft, and Visser}]{carter2023meaningful}
Owen B.~J. Carter, Shayne Loft, and Troy A.~W. Visser. 2023.
\newblock \href {https://doi.org/10.1177/00187208231218156} {Meaningful communication but not superficial anthropomorphism facilitates human-automation trust calibration: The human-automation trust expectation model (hatem)}.
\newblock \emph{Human Factors}, 66(11):2485--2502.

\bibitem[{Chattopadhyay and MacDorman(2017)}]{chattopadhyay2017perceptual}
Debaleena Chattopadhyay and Karl~F. MacDorman. 2017.
\newblock Perceptual expectations and the uncanny valley.
\newblock \url{https://citeseerx.ist.psu.edu/document?repid=rep1&type=pdf&doi=300d2231851a3d86939965ae1a51f2b2234056ee}.

\bibitem[{Chen et~al.(2024)Chen, Wang, Xu, Yuan, Zhang, Shi, Xie, Li, Yang, Zhu, Chen, Li, Chen, Hu, Wu, Ren, Fu, and Xiao}]{chen2024personapersonalizationsurveyroleplaying}
Jiangjie Chen, Xintao Wang, Rui Xu, Siyu Yuan, Yikai Zhang, Wei Shi, Jian Xie, Shuang Li, Ruihan Yang, Tinghui Zhu, Aili Chen, Nianqi Li, Lida Chen, Caiyu Hu, Siye Wu, Scott Ren, Ziquan Fu, and Yanghua Xiao. 2024.
\newblock \href {https://arxiv.org/abs/2404.18231} {From persona to personalization: A survey on role-playing language agents}.
\newblock \emph{Preprint}, arXiv:2404.18231.

\bibitem[{Cheng et~al.(2024)Cheng, Gligoric, Piccardi, and Jurafsky}]{cheng-etal-2024-anthroscore}
Myra Cheng, Kristina Gligoric, Tiziano Piccardi, and Dan Jurafsky. 2024.
\newblock \href {https://aclanthology.org/2024.eacl-long.49/} {{A}nthro{S}core: A computational linguistic measure of anthropomorphism}.
\newblock In \emph{Proceedings of the 18th Conference of the European Chapter of the Association for Computational Linguistics (Volume 1: Long Papers)}, pages 807--825, St. Julian{'}s, Malta. Association for Computational Linguistics.

\bibitem[{Cheng et~al.(2025)Cheng, Yu, and Jurafsky}]{cheng2025humtdumtmeasuringcontrolling}
Myra Cheng, Sunny Yu, and Dan Jurafsky. 2025.
\newblock \href {https://arxiv.org/abs/2502.13259} {Humt dumt: Measuring and controlling human-like language in llms}.
\newblock \emph{Preprint}, arXiv:2502.13259.

\bibitem[{Cohn et~al.(2024{\natexlab{a}})Cohn, Pushkarna, Olanubi, Moran, Padgett, Mengesha, and Heldreth}]{cohn2024believinganthropomorphismexaminingrole}
Michelle Cohn, Mahima Pushkarna, Gbolahan~O. Olanubi, Joseph~M. Moran, Daniel Padgett, Zion Mengesha, and Courtney Heldreth. 2024{\natexlab{a}}.
\newblock \href {https://arxiv.org/abs/2405.06079} {Believing anthropomorphism: Examining the role of anthropomorphic cues on trust in large language models}.
\newblock \emph{Preprint}, arXiv:2405.06079.

\bibitem[{Cohn et~al.(2024{\natexlab{b}})Cohn, Pushkarna, Olanubi, Moran, Padgett, Mengesha, and Heldreth}]{Cohn2024BelievingAE}
Michelle Cohn, Mahima Pushkarna, Gbolahan~O. Olanubi, Joseph~M. Moran, Daniel Padgett, Zion Mengesha, and Courtney Heldreth. 2024{\natexlab{b}}.
\newblock \href {https://api.semanticscholar.org/CorpusID:269740770} {Believing anthropomorphism: Examining the role of anthropomorphic cues on trust in large language models}.
\newblock \emph{Extended Abstracts of the CHI Conference on Human Factors in Computing Systems}.

\bibitem[{Coricelli(2005)}]{coricelli2005two}
Giorgio Coricelli. 2005.
\newblock Two-levels of mental states attribution: from automaticity to voluntariness.
\newblock \emph{Neuropsychologia}, 43(2):294--300.

\bibitem[{Crolic et~al.(2020)Crolic, Thomaz, Lamberton, and Stephen}]{crolic2020chatbot}
Camilla Crolic, Flavio Thomaz, Cait Lamberton, and Andrew~T Stephen. 2020.
\newblock \href {https://doi.org/10.1145/3319502.3374788} {Chatbot trust and anthropomorphism: The role of appearance and behavior}.
\newblock In \emph{Proceedings of the 2020 CHI Conference on Human Factors in Computing Systems}, pages 1--13.

\bibitem[{Crolic et~al.(2022)Crolic, Thomaz, Hadi, and Stephen}]{crolic2022blame}
Cammy Crolic, Felipe Thomaz, Rhonda Hadi, and Andrew~T. Stephen. 2022.
\newblock \href {https://doi.org/10.1177/00222429211045687} {Blame the bot: Anthropomorphism and anger in customer–chatbot interactions}.
\newblock \emph{Journal of Marketing}, 86(1):132--148.

\bibitem[{Culley and Madhavan(2013)}]{CULLEY2013577}
Kimberly~E. Culley and Poornima Madhavan. 2013.
\newblock \href {https://doi.org/10.1016/j.chb.2012.11.023} {A note of caution regarding anthropomorphism in hci agents}.
\newblock \emph{Computers in Human Behavior}, 29(3):577--579.

\bibitem[{Dai and Xiao(2025)}]{dai2025embracingcontradictiontheoreticalinconsistency}
Gordon Dai and Yunze Xiao. 2025.
\newblock \href {https://arxiv.org/abs/2505.18139} {Embracing contradiction: Theoretical inconsistency will not impede the road of building responsible ai systems}.
\newblock \emph{Preprint}, arXiv:2505.18139.

\bibitem[{Damholdt et~al.(2023)Damholdt, Quick, Seibt, Vestergaard, and Hansen}]{damholdt2023scoping}
Malene~Flensborg Damholdt, Oliver~Santiago Quick, Johanna Seibt, Christina Vestergaard, and Mads Hansen. 2023.
\newblock \href {https://doi.org/10.1007/s12369-023-01014-z} {A scoping review of hri research on 'anthropomorphism': Contributions to the method debate in hri}.
\newblock \emph{International Journal of Social Robotics}, 15(7):1203--1226.

\bibitem[{DeVrio et~al.(2025{\natexlab{a}})DeVrio, Cheng, Egede, Olteanu, and Blodgett}]{DeVrio_2025}
Alicia DeVrio, Myra Cheng, Lisa Egede, Alexandra Olteanu, and Su~Lin Blodgett. 2025{\natexlab{a}}.
\newblock \href {https://doi.org/10.1145/3706598.3714038} {A taxonomy of linguistic expressions that contribute to anthropomorphism of language technologies}.
\newblock In \emph{Proceedings of the 2025 CHI Conference on Human Factors in Computing Systems}, CHI ’25, page 1–18. ACM.

\bibitem[{DeVrio et~al.(2025{\natexlab{b}})DeVrio, Cheng, Egede, Olteanu, and Blodgett}]{Alice2025}
Alicia DeVrio, Myra Cheng, Lisa Egede, Alexandra Olteanu, and Su~Lin Blodgett. 2025{\natexlab{b}}.
\newblock \href {https://doi.org/10.1145/3706598.3714038} {A taxonomy of linguistic expressions that contribute to anthropomorphism of language technologies}.
\newblock In \emph{Proceedings of the 2025 CHI Conference on Human Factors in Computing Systems}, CHI '25, New York, NY, USA. Association for Computing Machinery.

\bibitem[{De Gennaro et~al.(2020)De Gennaro, Krumhuber, and Lucas}]{DeGennaro2020}
 Mauro De Gennaro, Eva G. Krumhuber, and  Gale M. Lucas. 2020.
\newblock \href {https://doi.org/10.3389/fpsyg.2019.03061} {Effectiveness of an empathic chatbot in combating adverse effects of social exclusion on mood}.
\newblock \emph{Frontiers in Psychology}, 10:3061.

\bibitem[{Don et~al.(1992)Don, Brennan, Laurel, and Shneiderman}]{don1992}
Abbe Don, Susan Brennan, Brenda Laurel, and Ben Shneiderman. 1992.
\newblock \href {https://doi.org/10.1145/142750.142760} {Anthropomorphism: from eliza to terminator 2}.
\newblock In \emph{Proceedings of the SIGCHI Conference on Human Factors in Computing Systems}, CHI '92, page 67–70, New York, NY, USA. Association for Computing Machinery.

\bibitem[{Duffau and Tree(2024)}]{Duffau2024ExpectingPP}
Elise Duffau and Jean E.~Fox Tree. 2024.
\newblock \href {https://api.semanticscholar.org/CorpusID:270935688} {Expecting politeness: perceptions of voice assistant politeness}.
\newblock \emph{Pers. Ubiquitous Comput.}, 28:907--929.

\bibitem[{Duffy(2003)}]{Duffy2003AnthropomorphismAT}
Brian~R. Duffy. 2003.
\newblock \href {https://api.semanticscholar.org/CorpusID:1959145} {Anthropomorphism and the social robot}.
\newblock \emph{Robotics Auton. Syst.}, 42:177--190.

\bibitem[{Ehrlich et~al.(2023)Ehrlich, Dean-Leon, Tacca, Armleder, Dimova-Edeleva, and Cheng}]{Ehrlich2023}
Stefan K. Ehrlich, Emmanuel Dean-Leon, Nicholas Tacca, Simon Armleder, Viktorija Dimova-Edeleva, and Gordon Cheng. 2023.
\newblock \href {https://doi.org/10.1371/journal.pone.0287958} {Human–robot collaborative task planning using anticipatory brain responses}.
\newblock \emph{PLOS ONE}, 18(7):e0287958.

\bibitem[{Epley et~al.(2008)Epley, Waytz, Akalis, and Cacioppo}]{epley2008when}
Nicholas Epley, Adam Waytz, Scott Akalis, and John~T Cacioppo. 2008.
\newblock \href {https://doi.org/10.1521/soco.2008.26.2.143} {When we need a human: Motivational determinants of anthropomorphism}.
\newblock \emph{Social Cognition}, 26(2):143--155.

\bibitem[{Epley et~al.(2007)Epley, Waytz, and Cacioppo}]{Epley2007OnSH}
Nicholas Epley, Adam Waytz, and John~T. Cacioppo. 2007.
\newblock \href {https://api.semanticscholar.org/CorpusID:6733517} {On seeing human: a three-factor theory of anthropomorphism.}
\newblock \emph{Psychological review}, 114 4:864--86.

\bibitem[{Eyssel and Hegel(2021)}]{eyssel2021trust}
Friederike Eyssel and Frank Hegel. 2021.
\newblock \href {https://doi.org/10.3389/frobt.2021.640444} {Can robots earn our trust the same way humans do? a systematic exploration of competence, warmth, and anthropomorphism as determinants of trust development in hri}.
\newblock \emph{Frontiers in Robotics and AI}, 8:640444.

\bibitem[{Eyssel et~al.(2015)Eyssel, Kuchenbrandt, Bobinger, de~Ruiter, and Hegel}]{eyssel2015cultural}
Friederike Eyssel, Dieta Kuchenbrandt, Sarah Bobinger, Laura de~Ruiter, and Frank Hegel. 2015.
\newblock \href {https://doi.org/10.3389/fpsyg.2015.00204} {Cultural differences in the evaluation of human-like robots}.
\newblock \emph{Frontiers in Psychology}, 6:204.

\bibitem[{Fong et~al.(2003)Fong, Nourbakhsh, and Dautenhahn}]{fong2003}
Terrence~W. Fong, Illah Nourbakhsh, and Kerstin Dautenhahn. 2003.
\newblock A survey of socially interactive robots.
\newblock \emph{Robotics and Autonomous Systems}, 42(3):143--166.

\bibitem[{Frazer(2022)}]{frazer2022experimental}
Rebecca Frazer. 2022.
\newblock \href {https://doi.org/10.1080/08934215.2022.2108472} {Experimental operationalizations of anthropomorphism in hci contexts: A scoping review}.
\newblock \emph{Communication Reports}, 35(3):123--136.

\bibitem[{Gillath et~al.(2021)Gillath, Ai, Branicky, Keshmiri, Davison, and Spaulding}]{gillath2021attachment}
Omri Gillath, Ting Ai, Michael~S Branicky, Shawn Keshmiri, Robert~B Davison, and Ryan Spaulding. 2021.
\newblock Attachment and trust in artificial intelligence.
\newblock \emph{Computers in Human Behavior}, 115:106607.

\bibitem[{Goyal et~al.(2024)Goyal, Chang, and Terry}]{goyal2024designing}
Nitesh Goyal, Minsuk Chang, and Michael Terry. 2024.
\newblock \href {https://arxiv.org/abs/2404.04289} {Designing for human-agent alignment: Understanding what humans want from their agents}.
\newblock \emph{arXiv preprint arXiv:2404.04289}.

\bibitem[{Guo et~al.(2025)Guo, Yang, Zhang, Song, Zhang, Xu, Zhu, Ma, Wang, Bi et~al.}]{guo2025deepseek}
Daya Guo, Dejian Yang, Haowei Zhang, Junxiao Song, Ruoyu Zhang, Runxin Xu, Qihao Zhu, Shirong Ma, Peiyi Wang, Xiao Bi, et~al. 2025.
\newblock Deepseek-r1: Incentivizing reasoning capability in llms via reinforcement learning.
\newblock \emph{arXiv preprint arXiv:2501.12948}.

\bibitem[{Guthrie(1997)}]{guthrie1997anthropomorphism}
Stewart~E. Guthrie. 1997.
\newblock Anthropomorphism: A definition and a theory.
\newblock In Robert~W. Mitchell, Nicholas~S. Thompson, and H.~Lyn Miles, editors, \emph{Anthropomorphism, Anecdotes, and Animals}, pages 50--58. State University of New York Press.

\bibitem[{Huang et~al.(2024{\natexlab{a}})Huang, LAM, Li, Ren, Wang, Jiao, Tu, and Lyu}]{NEURIPS2024_b0049c3f}
Jen-Tse Huang, Man~Ho LAM, Eric~John Li, Shujie Ren, Wenxuan Wang, Wenxiang Jiao, Zhaopeng Tu, and Michael~R Lyu. 2024{\natexlab{a}}.
\newblock \href {https://proceedings.neurips.cc/paper_files/paper/2024/file/b0049c3f9c53fb06f674ae66c2cf2376-Paper-Conference.pdf} {Apathetic or empathetic? evaluating llms\textquotesingle emotional alignments with humans}.
\newblock In \emph{Advances in Neural Information Processing Systems}, volume~37, pages 97053--97087. Curran Associates, Inc.

\bibitem[{Huang et~al.(2023)Huang, Tao, Zhang, An, Jiang, Chen, Wu, and Feng}]{huang2023lawyerllamatechnicalreport}
Quzhe Huang, Mingxu Tao, Chen Zhang, Zhenwei An, Cong Jiang, Zhibin Chen, Zirui Wu, and Yansong Feng. 2023.
\newblock \href {https://arxiv.org/abs/2305.15062} {Lawyer llama technical report}.
\newblock \emph{Preprint}, arXiv:2305.15062.

\bibitem[{Huang et~al.(2024{\natexlab{b}})Huang, Lin, He, Huang, and Huang}]{10.1145/3613905.3650823}
Shih-Hong Huang, Ya-Fang Lin, Zeyu He, Chieh-Yang Huang, and Ting-Hao~Kenneth Huang. 2024{\natexlab{b}}.
\newblock \href {https://doi.org/10.1145/3613905.3650823} {How does conversation length impact user’s satisfaction? a case study of length-controlled conversations with llm-powered chatbots}.
\newblock In \emph{Extended Abstracts of the CHI Conference on Human Factors in Computing Systems}, CHI EA '24, New York, NY, USA. Association for Computing Machinery.

\bibitem[{Ibrahim et~al.(2025)Ibrahim, Akbulut, Elasmar, Rastogi, Kahng, Morris, McKee, Rieser, Shanahan, and Weidinger}]{ibrahim2025multiturnevaluationanthropomorphicbehaviours}
Lujain Ibrahim, Canfer Akbulut, Rasmi Elasmar, Charvi Rastogi, Minsuk Kahng, Meredith~Ringel Morris, Kevin~R. McKee, Verena Rieser, Murray Shanahan, and Laura Weidinger. 2025.
\newblock \href {https://arxiv.org/abs/2502.07077} {Multi-turn evaluation of anthropomorphic behaviours in large language models}.
\newblock \emph{Preprint}, arXiv:2502.07077.

\bibitem[{Jaidka et~al.(2024)Jaidka, Ahuja, and Ng}]{jaidka2024takes}
Kokil Jaidka, Hansin Ahuja, and Lynnette Hui~Xian Ng. 2024.
\newblock It takes two to negotiate: Modeling social exchange in online multiplayer games.
\newblock \emph{Proceedings of the ACM on Human-Computer Interaction}, 8(CSCW1):1--22.

\bibitem[{Jiang et~al.(2022)Jiang, Liu, and Brooks}]{jiang2022trust}
Kai Jiang, Hao Liu, and Rodney Brooks. 2022.
\newblock \href {https://arxiv.org/abs/2208.14637} {Trust-performance paradox in industrial hri: Empirical evidence from warehouse robotics}.
\newblock arXiv preprint arXiv:2208.14637.

\bibitem[{Kang et~al.(2025)Kang, Freitas~dos Santos, Ben~Moussa, and Magnenat-Thalmann}]{kang2025mitigating}
Hangyeol Kang, Thiago Freitas~dos Santos, Maher Ben~Moussa, and Nadia Magnenat-Thalmann. 2025.
\newblock \href {https://arxiv.org/abs/2503.16449} {Mitigating the uncanny valley effect in hyper-realistic robots: A student-centered study on llm-driven conversations}.
\newblock \emph{arXiv preprint arXiv:2503.16449}.

\bibitem[{Kennedy(1992)}]{kennedy1992new}
John~S. Kennedy. 1992.
\newblock \emph{The New Anthropomorphism}.
\newblock Cambridge University Press, Cambridge, UK.

\bibitem[{Kim et~al.(2024)Kim, Liao, Vorvoreanu, Ballard, and Vaughan}]{Kim2024}
Sunnie S. Y. Kim, Q. Vera Liao, Mihaela Vorvoreanu, Stephanie Ballard, and Jennifer W. Vaughan. 2024.
\newblock \href {https://doi.org/10.1145/3630106.3658941} {{“I’m Not Sure, But …”}: Examining the impact of large language models’ uncertainty expression on user reliance and trust}.
\newblock In \emph{Proceedings of the 2024 ACM Conference on Fairness, Accountability, and Transparency}.

\bibitem[{Kim and Sundar(2012)}]{Kim2012AnthropomorphismOC}
Youjeong Kim and S.~Shyam Sundar. 2012.
\newblock \href {https://api.semanticscholar.org/CorpusID:37697957} {Anthropomorphism of computers: Is it mindful or mindless?}
\newblock \emph{Comput. Hum. Behav.}, 28:241--250.

\bibitem[{King and Ohya(1996)}]{king1996}
William~Joseph King and Jun Ohya. 1996.
\newblock \href {https://doi.org/10.1145/257089.257326} {The representation of agents: anthropomorphism, agency, and intelligence}.
\newblock In \emph{Conference Companion on Human Factors in Computing Systems}, CHI '96, page 289–290, New York, NY, USA. Association for Computing Machinery.

\bibitem[{Kulms and Kopp(2016)}]{kulms2016anthropomorphic}
Philipp Kulms and Stefan Kopp. 2016.
\newblock Anthropomorphic design: A framework for human-like interaction.
\newblock \url{https://citeseerx.ist.psu.edu/document?repid=rep1&type=pdf&doi=7f5a2fa14e695b5de864d86b6ce92b3a3b538817}.

\bibitem[{Kulms and Kopp(2023)}]{kulms2023anthropomorphic}
Philipp Kulms and Stefan Kopp. 2023.
\newblock \href {https://doi.org/10.1177/21695067231196240} {Anthropomorphic design: A framework for human-like interaction}.
\newblock \emph{International Journal of Human-Computer Studies}, 170:102933.

\bibitem[{Lee and Hahn(2024)}]{Lee2024}
Inju Lee and Sowon Hahn. 2024.
\newblock \href {https://doi.org/10.3389/fpsyg.2024.1282036} {On the relationship between mind perception and social support of chatbots}.
\newblock \emph{Frontiers in Psychology}, 15:1282036.

\bibitem[{Li and Suh(2022)}]{Li2022Anthropomorphism}
Mengjun Li and Ayoung Suh. 2022.
\newblock \href {https://doi.org/10.1007/s12525-022-00591-7} {Anthropomorphism in {AI}-enabled technology: A literature review}.
\newblock \emph{Electronic Markets}, 32(4):2245--2275.

\bibitem[{Li and Sung(2021)}]{LI2021106680}
Xinge Li and Yongjun Sung. 2021.
\newblock \href {https://doi.org/10.1016/j.chb.2021.106680} {Anthropomorphism brings us closer: The mediating role of psychological distance in user–ai assistant interactions}.
\newblock \emph{Computers in Human Behavior}, 118:106680.

\bibitem[{Liu et~al.(2024{\natexlab{a}})Liu, Diab, and Fried}]{liu-etal-2024-evaluating-large}
Andy Liu, Mona Diab, and Daniel Fried. 2024{\natexlab{a}}.
\newblock \href {https://doi.org/10.18653/v1/2024.findings-acl.586} {Evaluating large language model biases in persona-steered generation}.
\newblock In \emph{Findings of the Association for Computational Linguistics: ACL 2024}, pages 9832--9850, Bangkok, Thailand. Association for Computational Linguistics.

\bibitem[{Liu et~al.(2025)Liu, Song, Xiao, Zheng, Tjuatja, Borg, Diab, and Sap}]{liu2025syntheticsocraticdebatesexamining}
Jiarui Liu, Yueqi Song, Yunze Xiao, Mingqian Zheng, Lindia Tjuatja, Jana~Schaich Borg, Mona Diab, and Maarten Sap. 2025.
\newblock \href {https://arxiv.org/abs/2506.12657} {Synthetic socratic debates: Examining persona effects on moral decision and persuasion dynamics}.
\newblock \emph{Preprint}, arXiv:2506.12657.

\bibitem[{Liu and Tao(2022)}]{Liu2022TheRO}
Kaifeng Liu and Da~Tao. 2022.
\newblock \href {https://api.semanticscholar.org/CorpusID:239118721} {The roles of trust, personalization, loss of privacy, and anthropomorphism in public acceptance of smart healthcare services}.
\newblock \emph{Comput. Hum. Behav.}, 127:107026.

\bibitem[{Liu et~al.(2024{\natexlab{b}})Liu, Giorgi, Aich, Lahnala, Curtis, Ungar, and Sedoc}]{Liu2024}
 Tingting Liu,  Salvatore Giorgi,  Ankit Aich, Allison Lahnala,  Brenda Curtis,  Lyle Ungar, and  Jo{\~a}o Sedoc. 2024{\natexlab{b}}.
\newblock The illusion of empathy: How ai chatbots shape conversation perception.
\newblock \emph{arXiv preprint arXiv:2411.12877}.

\bibitem[{Luo et~al.(2025)Luo, Yang, Cai, Zhang, and Zheng}]{Luo2025}
Zhen Luo, Yixuan Yang, Chang Cai, Yanfu Zhang, and  Feng Zheng. 2025.
\newblock Roboreflect: Robotic reflective reasoning for grasping ambiguous‑condition objects.
\newblock \emph{arXiv preprint arXiv:2501.09307}.

\bibitem[{Ma et~al.(2024)Ma, Hu, Li, Wang, Liu, and Cheong}]{ma2024studentsexpertsnewai}
Yiping Ma, Shiyu Hu, Xuchen Li, Yipei Wang, Shiqing Liu, and Kang~Hao Cheong. 2024.
\newblock \href {https://arxiv.org/abs/2410.15701} {Students rather than experts: A new ai for education pipeline to model more human-like and personalised early adolescences}.
\newblock \emph{Preprint}, arXiv:2410.15701.

\bibitem[{Marchesi et~al.(2022)Marchesi, Ricci, Cavallo, and Esposito}]{marchesi2022mental}
Stefano Marchesi, Elisa Ricci, Filippo Cavallo, and Roberto Esposito. 2022.
\newblock \href {https://doi.org/10.1145/3526112} {Mental state attribution to robots: A systematic review of anthropomorphism in human–robot interaction}.
\newblock \emph{ACM Computing Surveys (CSUR)}, 55(3):1--36.

\bibitem[{Mathur et~al.(2025)Mathur, Qian, Liang, and Morency}]{mathur2025socialgenomegroundedsocial}
Leena Mathur, Marian Qian, Paul~Pu Liang, and Louis-Philippe Morency. 2025.
\newblock \href {https://arxiv.org/abs/2502.15109} {Social genome: Grounded social reasoning abilities of multimodal models}.
\newblock \emph{Preprint}, arXiv:2502.15109.

\bibitem[{Mireshghallah et~al.(2024)Mireshghallah, Antoniak, More, Choi, and Farnadi}]{mireshghallah2024trustbotdiscoveringpersonal}
Niloofar Mireshghallah, Maria Antoniak, Yash More, Yejin Choi, and Golnoosh Farnadi. 2024.
\newblock \href {https://arxiv.org/abs/2407.11438} {Trust no bot: Discovering personal disclosures in human-llm conversations in the wild}.
\newblock \emph{Preprint}, arXiv:2407.11438.

\bibitem[{Mitchell et~al.(1997)Mitchell, Thompson, and Miles}]{mitchell1997anthropomorphism}
Robert~W. Mitchell, Nicholas~S. Thompson, and H.~Lyn Miles, editors. 1997.
\newblock \emph{Anthropomorphism, Anecdotes, and Animals}.
\newblock State University of New York Press.

\bibitem[{Moore and Zhang(2024)}]{moore2024uncanny}
Erin Moore and Mingyu Zhang. 2024.
\newblock \href {https://doi.org/10.1016/j.chb.2024.107275} {Is the uncanny valley a myth? a meta-analysis of 127 empirical studies}.
\newblock \emph{Computers in Human Behavior}, 150:107275.

\bibitem[{Mori et~al.(2012)Mori, MacDorman, and Kageki}]{6213238}
Masahiro Mori, Karl~F. MacDorman, and Norri Kageki. 2012.
\newblock \href {https://doi.org/10.1109/MRA.2012.2192811} {The uncanny valley [from the field]}.
\newblock \emph{IEEE Robotics \& Automation Magazine}, 19(2):98--100.

\bibitem[{Moussawi et~al.(2020)Moussawi, Koufaris, and Benbunan-Fich}]{Moussawi2020HowPO}
Sara Moussawi, Marios Koufaris, and Raquel Benbunan-Fich. 2020.
\newblock \href {https://api.semanticscholar.org/CorpusID:216194438} {How perceptions of intelligence and anthropomorphism affect adoption of personal intelligent agents}.
\newblock \emph{Electronic Markets}, 31:343 -- 364.

\bibitem[{Nass and Moon(2000)}]{NassMoon2000}
Clifford Nass and Youngme Moon. 2000.
\newblock Machines and mindlessness: Social responses to computers.
\newblock In \emph{Journal of Social Issues}, volume~56, pages 81--103.

\bibitem[{Nass et~al.(1997)Nass, Moon, Green, and Reaves}]{nass1997anthropomorphism}
Clifford Nass, Youngme Moon, Nancy Green, and Byron Reaves. 1997.
\newblock \href {https://doi.org/10.1145/257089.257326} {Anthropomorphism, agency, and ethopoeia: Computers as social actors}.
\newblock In \emph{Proceedings of the SIGCHI Conference on Human Factors in Computing Systems}, pages 228--229.

\bibitem[{Nass et~al.(1994)Nass, Steuer, and Tauber}]{nass1994}
Clifford Nass, Jonathan Steuer, and Ellen~R. Tauber. 1994.
\newblock \href {https://doi.org/10.1145/191666.191703} {Computers are social actors}.
\newblock In \emph{Proceedings of the SIGCHI Conference on Human Factors in Computing Systems}, CHI '94, page 72–78, New York, NY, USA. Association for Computing Machinery.

\bibitem[{Natarajan and Gombolay(2020)}]{manisha2020}
Manisha Natarajan and Matthew Gombolay. 2020.
\newblock \href {https://doi.org/10.1145/3319502.3374839} {Effects of anthropomorphism and accountability on trust in human robot interaction}.
\newblock In \emph{Proceedings of the 2020 ACM/IEEE International Conference on Human-Robot Interaction}, HRI '20, page 33–42, New York, NY, USA. Association for Computing Machinery.

\bibitem[{Ng et~al.(2025)Ng, Kang, and Carley}]{ng2025aurasightgeneratingrealisticsocial}
Lynnette Hui~Xian Ng, Bianca N.~Y. Kang, and Kathleen~M. Carley. 2025.
\newblock \href {https://arxiv.org/abs/2509.08927} {Aurasight: Generating realistic social media data}.
\newblock \emph{Preprint}, arXiv:2509.08927.

\bibitem[{Nighojkar et~al.(2025)Nighojkar, Moydinboyev, Duong, and Licato}]{nighojkar2025givingaipersonalitiesleads}
Animesh Nighojkar, Bekhzodbek Moydinboyev, My~Duong, and John Licato. 2025.
\newblock \href {https://arxiv.org/abs/2502.14155} {Giving ai personalities leads to more human-like reasoning}.
\newblock \emph{Preprint}, arXiv:2502.14155.

\bibitem[{Olteanu et~al.(2025)Olteanu, Barocas, Blodgett, Egede, DeVrio, and Cheng}]{olteanu2025aiautomatonsaisystems}
Alexandra Olteanu, Solon Barocas, Su~Lin Blodgett, Lisa Egede, Alicia DeVrio, and Myra Cheng. 2025.
\newblock \href {https://arxiv.org/abs/2503.02250} {Ai automatons: Ai systems intended to imitate humans}.
\newblock \emph{Preprint}, arXiv:2503.02250.

\bibitem[{Paivio(1978)}]{paivio1978mental}
Allan Paivio. 1978.
\newblock Mental comparisons involving abstract attributes.
\newblock \emph{Memory \& Cognition}, 6(3):199--208.

\bibitem[{Park and Whang(2022)}]{Park2022}
Sung Park and Mincheol Whang. 2022.
\newblock \href {https://doi.org/10.3390/ijerph19031889} {Empathy in human–robot interaction: Designing for social robots}.
\newblock \emph{International Journal of Environmental Research and Public Health}, 19(3):1889.

\bibitem[{Parsons(2023)}]{parsons2023ethics}
Thomas Parsons. 2023.
\newblock \href {https://philarchive.org/archive/PARTEO-58} {The ethics of ai at the intersection of transgender identity and neurodiversity}.
\newblock \emph{Philosophical Archive}.

\bibitem[{Pennycook and Rand(2020)}]{pennycook2020falls}
Gordon Pennycook and David~G Rand. 2020.
\newblock Who falls for fake news? the roles of bullshit receptivity, overclaiming, familiarity, and analytic thinking.
\newblock \emph{Journal of personality}, 88(2):185--200.

\bibitem[{Peter et~al.(2025)Peter, Riemer, and West}]{doi:10.1073/pnas.2415898122}
Sandra Peter, Kai Riemer, and Jevin~D. West. 2025.
\newblock \href {https://doi.org/10.1073/pnas.2415898122} {The benefits and dangers of anthropomorphic conversational agents}.
\newblock \emph{Proceedings of the National Academy of Sciences}, 122(22):e2415898122.

\bibitem[{Pinch and Bijker(1984)}]{pinch1984social}
Trevor~J. Pinch and Wiebe~E. Bijker. 1984.
\newblock \href {https://doi.org/10.1177/030631284014003004} {The social construction of facts and artefacts: Or how the sociology of science and the sociology of technology might benefit each other}.
\newblock \emph{Social Studies of Science}, 14(3):399--441.

\bibitem[{Placani(2024)}]{placani2024anthropomorphism}
Adriana Placani. 2024.
\newblock \href {https://doi.org/10.1007/s43681-024-00419-4} {Anthropomorphism in {AI}: hype and fallacy}.
\newblock \emph{AI and Ethics}, 4.

\bibitem[{Reeves and Nass(1996)}]{Reeves1996}
Byron Reeves and Clifford Nass. 1996.
\newblock \emph{The Media Equation: How People Treat Computers, Television, and New Media Like Real People and Places}.
\newblock Cambridge University Press.

\bibitem[{Reicherts et~al.(2021)Reicherts, Zargham, Bonfert, Rogers, and Malaka}]{reicherts2021interrupt}
Leon Reicherts, Neda Zargham, Michael Bonfert, Yvonne Rogers, and Rainer Malaka. 2021.
\newblock \href {https://doi.org/10.1145/3469595.3469629} {May i interrupt? diverging opinions on proactive smart speakers}.
\newblock In \emph{Proceedings of the 3rd Conference on Conversational User Interfaces (CUI '21)}, pages 1--10, New York, NY, USA. Association for Computing Machinery.

\bibitem[{Sah and Wei(2015)}]{sah2015effects}
Y.~Sah and Y.~Wei. 2015.
\newblock \href {https://doi.org/10.1016/j.chb.2014.12.068} {Effects of visual and linguistic anthropomorphic cues on social perception, self-awareness, and information disclosure in a health website}.
\newblock \emph{Computers in Human Behavior}, 45:392--401.

\bibitem[{Salem et~al.(2013)Salem, Eyssel, Rohlfing, Kopp, and Joublin}]{Salem2013ToEI}
Maha Salem, Friederike Eyssel, Katharina~J. Rohlfing, Stefan Kopp, and Frank Joublin. 2013.
\newblock \href {https://api.semanticscholar.org/CorpusID:10900363} {To err is human(-like): Effects of robot gesture on perceived anthropomorphism and likability}.
\newblock \emph{International Journal of Social Robotics}, 5:313 -- 323.

\bibitem[{Schramowski et~al.(2021)Schramowski, Turan, Andersen, Rothkopf, and Kersting}]{schramowski2021large}
Patrick Schramowski, Cigdem Turan, Nico Andersen, Constantin~A Rothkopf, and Kristian Kersting. 2021.
\newblock Large pre-trained language models contain human-like biases of what is right and wrong to do.
\newblock \emph{arXiv preprint arXiv:2103.11790}.

\bibitem[{Seitz(2024)}]{Seitz2024}
Lennart Seitz. 2024.
\newblock \href {https://doi.org/10.1016/j.chbah.2024.100067} {Artificial empathy in healthcare chatbots: Does it feel authentic?}
\newblock \emph{Computers in Human Behavior: Artificial Humans}, 2:100067.

\bibitem[{Shiomi et~al.(2024)Shiomi, Hirayama, Kimoto, Iio, and Shimohara}]{shiomi2024bowing}
Masahiro Shiomi, Taichi Hirayama, Mitsuhiko Kimoto, Takamasa Iio, and Katsunori Shimohara. 2024.
\newblock Modeling of bowing behaviors in apology based on severity.
\newblock \emph{IEEE Robotics and Automation Letters}, 9(11):10169--10176.

\bibitem[{Stiber et~al.(2025)Stiber, Taylor, and Huang}]{Stiber2025}
Maia Stiber, Russell Taylor, and Chien‑Ming Huang. 2025.
\newblock Robot error awareness through human reactions: Implementation, evaluation, and recommendations.
\newblock \emph{arXiv preprint arXiv:2501.05723}.

\bibitem[{Tejeda et~al.(2025)}]{tejeda2025what}
Heliodoro Tejeda et~al. 2025.
\newblock \href {https://doi.org/10.1038/s42256-024-00976-7} {What large language models know and what people think they know}.
\newblock \emph{Nature Machine Intelligence}.

\bibitem[{tse Huang et~al.(2024{\natexlab{a}})tse Huang, Lam, Li, Ren, Wang, Jiao, Tu, and Lyu}]{huang2024emotionallynumbempatheticevaluating}
Jen tse Huang, Man~Ho Lam, Eric~John Li, Shujie Ren, Wenxuan Wang, Wenxiang Jiao, Zhaopeng Tu, and Michael~R. Lyu. 2024{\natexlab{a}}.
\newblock \href {https://arxiv.org/abs/2308.03656} {Emotionally numb or empathetic? evaluating how llms feel using emotionbench}.
\newblock \emph{Preprint}, arXiv:2308.03656.

\bibitem[{tse Huang et~al.(2024{\natexlab{b}})tse Huang, Wang, Li, Lam, Ren, Yuan, Jiao, Tu, and Lyu}]{huang2024chatgptbenchmarkingllmspsychological}
Jen tse Huang, Wenxuan Wang, Eric~John Li, Man~Ho Lam, Shujie Ren, Youliang Yuan, Wenxiang Jiao, Zhaopeng Tu, and Michael~R. Lyu. 2024{\natexlab{b}}.
\newblock \href {https://arxiv.org/abs/2310.01386} {Who is chatgpt? benchmarking llms' psychological portrayal using psychobench}.
\newblock \emph{Preprint}, arXiv:2310.01386.

\bibitem[{Tsoukalas et~al.(2024)Tsoukalas, Lee, Jennings, Xin, Ding, Jennings, Thakur, and Chaudhuri}]{tsoukalas2024putnambenchevaluatingneuraltheoremprovers}
George Tsoukalas, Jasper Lee, John Jennings, Jimmy Xin, Michelle Ding, Michael Jennings, Amitayush Thakur, and Swarat Chaudhuri. 2024.
\newblock \href {https://arxiv.org/abs/2407.11214} {Putnambench: Evaluating neural theorem-provers on the putnam mathematical competition}.
\newblock \emph{Preprint}, arXiv:2407.11214.

\bibitem[{Tu and Lee(2023)}]{tu2023effects}
Jiexin Tu and Joonhwan Lee. 2023.
\newblock \href {https://doi.org/10.1016/j.jretconser.2023.103446} {Effects of anthropomorphic design cues of chatbots on users' perception and visual behaviors}.
\newblock \emph{Journal of Retailing and Consumer Services}, 75:103446.

\bibitem[{Turing(1950)}]{turing1950computing}
Alan~M. Turing. 1950.
\newblock \href {https://doi.org/10.1093/mind/LIX.236.433} {Computing machinery and intelligence}.
\newblock \emph{Mind}, 59(236):433--460.

\bibitem[{Urakami and Seaborn(2022)}]{urakami2022nonverbal}
Jacqueline Urakami and Katie Seaborn. 2022.
\newblock \href {https://doi.org/10.1145/3570169} {Nonverbal cues in human–robot interaction: A communication studies perspective}.
\newblock \emph{ACM Transactions on Human-Robot Interaction}, 11(4):1--21.

\bibitem[{Urquiza-Haas and Kotrschal(2015)}]{urquiza2015mind}
Nayeli Urquiza-Haas and Kurt Kotrschal. 2015.
\newblock \href {https://doi.org/10.1016/j.anbehav.2015.08.011} {The mind behind anthropomorphic thinking: Attribution of mental states to other species}.
\newblock \emph{Animal Behaviour}, 109:167--176.

\bibitem[{Velner et~al.(2021)Velner, Truong, and Evers}]{velner2021speaking}
Ella Velner, Khiet~P. Truong, and Vanessa Evers. 2021.
\newblock \href {https://arxiv.org/abs/2104.05340} {Speaking of trust -- speech as a measure of trust}.
\newblock \emph{arXiv preprint arXiv:2104.05340}.

\bibitem[{Wang et~al.(2024{\natexlab{a}})Wang, Milani, Chiu, Zhi, Eack, Labrum, Murphy, Jones, Hardy, Shen, Fang, and Chen}]{wang-etal-2024-patient}
Ruiyi Wang, Stephanie Milani, Jamie~C. Chiu, Jiayin Zhi, Shaun~M. Eack, Travis Labrum, Samuel~M Murphy, Nev Jones, Kate~V Hardy, Hong Shen, Fei Fang, and Zhiyu Chen. 2024{\natexlab{a}}.
\newblock \href {https://doi.org/10.18653/v1/2024.emnlp-main.711} {{PATIENT}-$\psi$: Using large language models to simulate patients for training mental health professionals}.
\newblock In \emph{Proceedings of the 2024 Conference on Empirical Methods in Natural Language Processing}, pages 12772--12797, Miami, Florida, USA. Association for Computational Linguistics.

\bibitem[{Wang et~al.(2024{\natexlab{b}})Wang, Tao, Fang, Wang, Wang, Jiang, and Zhou}]{wang2024aipersonalifelongpersonalization}
Tiannan Wang, Meiling Tao, Ruoyu Fang, Huilin Wang, Shuai Wang, Yuchen~Eleanor Jiang, and Wangchunshu Zhou. 2024{\natexlab{b}}.
\newblock \href {https://arxiv.org/abs/2412.13103} {Ai persona: Towards life-long personalization of llms}.
\newblock \emph{Preprint}, arXiv:2412.13103.

\bibitem[{Wang et~al.(2024{\natexlab{c}})Wang, Xiao, Huang, Yuan, Xu, Guo, Tu, Fei, Leng, Wang, Chen, Li, and Xiao}]{wang-etal-2024-incharacter}
Xintao Wang, Yunze Xiao, Jen-tse Huang, Siyu Yuan, Rui Xu, Haoran Guo, Quan Tu, Yaying Fei, Ziang Leng, Wei Wang, Jiangjie Chen, Cheng Li, and Yanghua Xiao. 2024{\natexlab{c}}.
\newblock \href {https://doi.org/10.18653/v1/2024.acl-long.102} {{I}n{C}haracter: Evaluating personality fidelity in role-playing agents through psychological interviews}.
\newblock In \emph{Proceedings of the 62nd Annual Meeting of the Association for Computational Linguistics (Volume 1: Long Papers)}, pages 1840--1873, Bangkok, Thailand. Association for Computational Linguistics.

\bibitem[{Waytz et~al.(2014)Waytz, Heafner, and Epley}]{Waytz2014TheMI}
Adam Waytz, Joy Heafner, and Nicholas Epley. 2014.
\newblock \href {https://api.semanticscholar.org/CorpusID:1302225} {The mind in the machine: Anthropomorphism increases trust in an autonomous vehicle}.
\newblock \emph{Journal of Experimental Social Psychology}, 52:113--117.

\bibitem[{Weizenbaum(1966)}]{weizenbaum1966eliza}
Joseph Weizenbaum. 1966.
\newblock \href {https://doi.org/10.1145/365153.365168} {Eliza—a computer program for the study of natural language communication between man and machine}.
\newblock \emph{Communications of the ACM}, 9(1):36--45.

\bibitem[{Wellman(1990)}]{wellman1990child}
Henry~M. Wellman. 1990.
\newblock \emph{The Child's Theory of Mind}.
\newblock MIT Press, Cambridge, MA.

\bibitem[{Wen et~al.(2024)Wen, Norel, Liu, Stappenbeck, Zulkernine, and Chen}]{Wen_2024}
Bo~Wen, Raquel Norel, Julia Liu, Thaddeus Stappenbeck, Farhana Zulkernine, and Huamin Chen. 2024.
\newblock \href {https://doi.org/10.1109/icdh62654.2024.00027} {Leveraging large language models for patient engagement: The power of conversational ai in digital health}.
\newblock In \emph{2024 IEEE International Conference on Digital Health (ICDH)}, page 104–113. IEEE.

\bibitem[{Wu et~al.(2024)Wu, Cachia, Han, Yao, Xie, Zhao, and Wang}]{wu2024ilikesunniei}
Siyi Wu, Julie Y.~A. Cachia, Feixue Han, Bingsheng Yao, Tianyi Xie, Xuan Zhao, and Dakuo Wang. 2024.
\newblock \href {https://arxiv.org/abs/2405.13803} {"i like sunnie more than i expected!": Exploring user expectation and perception of an anthropomorphic llm-based conversational agent for well-being support}.
\newblock \emph{Preprint}, arXiv:2405.13803.

\bibitem[{Xie et~al.(2024)Xie, Zhang, Chen, Zhu, Lou, Tian, Xiao, and Su}]{10.5555/3692070.3694316}
Jian Xie, Kai Zhang, Jiangjie Chen, Tinghui Zhu, Renze Lou, Yuandong Tian, Yanghua Xiao, and Yu~Su. 2024.
\newblock Travelplanner: a benchmark for real-world planning with language agents.
\newblock In \emph{Proceedings of the 41st International Conference on Machine Learning}, ICML'24. JMLR.org.

\bibitem[{Xie et~al.(2025)Xie, Min, Ji, Yang, Zhang, Xu, Bajaj, Salakhutdinov, Johnson-Roberson, and Bisk}]{xie2025embodiedraggeneralnonparametricembodied}
Quanting Xie, So~Yeon Min, Pengliang Ji, Yue Yang, Tianyi Zhang, Kedi Xu, Aarav Bajaj, Ruslan Salakhutdinov, Matthew Johnson-Roberson, and Yonatan Bisk. 2025.
\newblock \href {https://arxiv.org/abs/2409.18313} {Embodied-rag: General non-parametric embodied memory for retrieval and generation}.
\newblock \emph{Preprint}, arXiv:2409.18313.

\bibitem[{Xu et~al.(2025)Xu, Wang, Wang, Lu, Tan, Chu, and Xu}]{xu2025guessithinkingbenchmark}
Rui Xu, MingYu Wang, XinTao Wang, Dakuan Lu, Xiaoyu Tan, Wei Chu, and Yinghui Xu. 2025.
\newblock \href {https://arxiv.org/abs/2503.08193} {Guess what i am thinking: A benchmark for inner thought reasoning of role-playing language agents}.
\newblock \emph{Preprint}, arXiv:2503.08193.

\bibitem[{Yang et~al.(2025)Yang, Wu, Chen, Xiao, Yang, Wong, and Wang}]{yang2025ahamoments}
Shu Yang, Junchao Wu, Xin Chen, Yunze Xiao, Xinyi Yang, Derek~F Wong, and Di~Wang. 2025.
\newblock \href {https://arxiv.org/abs/2504.02956} {Understanding aha moments: From external observations to internal mechanisms}.
\newblock \emph{arXiv preprint arXiv:2504.02956}.

\bibitem[{Yang and Xie(2024)}]{yang2024}
Wenjing Yang and  Yunhui Xie. 2024.
\newblock \href {https://doi.org/10.1016/j.chbah.2024.100049} {Can robots elicit empathy? the effects of social robots’ appearance on emotional contagion}.
\newblock \emph{Computers in Human Behavior: Artificial Humans}, 2:100049.

\bibitem[{Yu(2011)}]{yu2011indirectness}
Myeonghee Yu. 2011.
\newblock Indirectness and politeness in english, hebrew, and korean requests.
\newblock \emph{Journal of Pragmatics}, 43(1):1--17.

\bibitem[{Yu and Park(2023)}]{yu2023cognitive}
X.~Yu and E.~Park. 2023.
\newblock \href {https://doi.org/10.3389/fpsyg.2023.11635073} {Cognitive dissonance in human–ai interaction: Implications for anthropomorphic design}.
\newblock \emph{Frontiers in Psychology}, 14:11635073.

\bibitem[{Yue(2025)}]{yue2025surveylargelanguagemodel}
Murong Yue. 2025.
\newblock \href {https://arxiv.org/abs/2503.19213} {A survey of large language model agents for question answering}.
\newblock \emph{Preprint}, arXiv:2503.19213.

\bibitem[{Zhou et~al.(2025)Zhou, Hwang, Ren, Dziri, Jurafsky, and Sap}]{zhou2025relai}
Kaitlyn Zhou, Jena~D. Hwang, Xiang Ren, Nouha Dziri, Dan Jurafsky, and Maarten Sap. 2025.
\newblock \href {https://arxiv.org/abs/2407.07950} {Rel-a.i.: An interaction-centered approach to measuring human-lm reliance}.
\newblock In \emph{NAACL}.

\bibitem[{Zhou et~al.(2022)Zhou, Sisman, Rana, Schuller, and Li}]{zhou2022speechsynthesismixedemotions}
Kun Zhou, Berrak Sisman, Rajib Rana, B.~W. Schuller, and Haizhou Li. 2022.
\newblock \href {https://arxiv.org/abs/2208.05890} {Speech synthesis with mixed emotions}.
\newblock \emph{Preprint}, arXiv:2208.05890.

\bibitem[{Z{\l}otowski et~al.(2015)Z{\l}otowski, Proudfoot, Yogeeswaran, and Bartneck}]{zlotowski2015anthropomorphism}
Jakub Z{\l}otowski, Diane Proudfoot, Kumar Yogeeswaran, and Christoph Bartneck. 2015.
\newblock Anthropomorphism: opportunities and challenges in human--robot interaction.
\newblock \emph{International journal of social robotics}, 7:347--360.

\end{thebibliography}
\newpage
\newpage

\appendix
\section{Definition Source Selection and Periodization}
\label{sec:app_definitions}

To conduct our diachronic synthesis of definitions of anthropomorphism, we implemented a structured review pipeline, combining historical periodization with a reproducible literature selection process. Our approach is rooted in identifying conceptual inflection points across three interdisciplinary domains: Robotics, Human-Agent Interaction (HAI), and Information Science.

\subsection{Periodization Criteria}

We segmented the literature into three eras, each reflecting a dominant technological paradigm or theoretical orientation. The division is grounded in historical developments and citation patterns.

\begin{itemize}
    \item \textbf{Era I (Pre-2000): Foundational and Theoretical Origins.}  
    \textit{Rationale:} This period includes foundational philosophical, psychological, and early cognitive science works that established core concepts around anthropomorphism (e.g., Piaget, Guthrie, Dennett). Robotics and AI remained largely symbolic or rule-based.  
    \textit{Semantic Scholar Filter:} Publication date \texttt{$\leq$ 1999}.  
    \textit{Selection Criteria:} Highest-cited conceptual papers containing explicit definitions or theoretical characterizations of anthropomorphism. Priority was given to publications in journals of psychology, HCI, and philosophy.

    \item \textbf{Era II (2000–2015): Embodied Agents and HRI Emergence.}  
    \textit{Rationale:} Marked by the rise of embodied social robots, virtual agents, and the first wave of HRI studies. Increasing emphasis on user interaction, social cues, and design frameworks.  
    \textit{Semantic Scholar Filter:} Publication date \texttt{2000–2015}.  
    \textit{Selection Criteria:} Definitions from empirical studies or design frameworks frequently cited in HRI, Social Robotics, or Computer-Supported Cooperative Work (CSCW).

    \item \textbf{Era III (2016–present): LLMs, Dynamic Interaction and Cultural Reflection.}  
    \textit{Rationale:} The era of deep learning, generative AI and renewed scrutiny of anthropomorphism in black-box systems. Includes work on LLM, moral expectations, and cross-cultural design.  
    \textit{Semantic Scholar Filter:} Publication date \texttt{$\geq$ 2016}.  
    \textit{Selection Criteria:} Highly cited or thematically central papers addressing anthropomorphism in large-scale language models, explainable AI, or global HAI. Definitions framed within empirical evaluation or ethical critique were prioritized.
\end{itemize}

\subsection{Method of Source Identification}

We queried \texttt{Semantic Scholar} using the keyword \textbf{``anthropomorphism"} and filtered the results by publication date for each of the three eras defined above. For each period, we extracted the top 20 articles most cited and performed a full text review to identify passages that provided formal definitions or operationalizations of anthropomorphism. Where multiple definitions were provided, we selected the most central or frequently cited variant. Cross-referencing was performed with Google Scholar and Scopus to confirm the citation patterns and disciplinary relevance.

\subsection{Final Corpus Composition}

In total, \textbf{33 unique definitions} were retained, covering HCI, HRI, and Information Science. These are presented chronologically in \autoref{tab:def_pre2000},\autoref{tab:def_2000to2015} and \autoref{tab:def_2016to2025}, along with the citation context and disciplinary affiliation. This curated list underpins the diachronic analysis presented in the main text.

\begin{table*}[t]
\centering
\begin{tabular}{p{10cm}|p{3cm}}
\hline
\textbf{Definition} & \textbf{Reference} \\
\hline
Anthropomorphism is the ascription of human characteristics to non-human entities. &  \cite{caporael1986anthropomorphism}\\ \hline
Anthropomorphism – [means] attributing human characteristics to non-human entities. &  \cite{burghardt1991cognitive}\\\hline
{[}Anthropomorphic thinking is] simply built into us (i.e., an innate tendency of humans). &  \cite{kennedy1992new}\\\hline
assigning human characteristics to the computer&  \cite{don1992}\\\hline
It is a sincere, conscious belief that [a computer or robot] is human and/or deserving of human attributions. &  \cite{nass1994}\\\hline
...the anthropomorphic representation allows for a rich
set of easily identifiable behaviors and for social interaction.&  \cite{king1996}\\\hline
Anthropomorphism is a pervasive, perhaps universal, way of thinking. &  \cite{boyer1996what}\\\hline
People treat communication media as if they were human. &  \cite{Reeves1996}\\\hline
"People respond socially and naturally to media even though they believe it is not reasonable to do so …'' &  \cite{Reeves1996}\\\hline
Anthropomorphism…[is] the attribution of human characteristics to non-human things or events. &  \cite{guthrie1997anthropomorphism}\\\hline
It is the universal human tendency to ascribe human physical and mental characteristics to non-human entities, objects and events. &  \cite{mitchell1997anthropomorphism}\\
\hline
\end{tabular}
\caption{Literature definitions of anthropomorphism across Robotics and Human–Computer/AI-Interaction domains before 2000, ordered chronologically.}
\label{tab:def_pre2000}
\end{table*}

\begin{table*}[t]
\centering
\begin{tabular}{p{10cm}|p{3cm}}
\hline
\textbf{Definition} & \textbf{Reference} \\
\hline
Anthropomorphism, the assignment of human traits and characteristics to computers.&  \cite{NassMoon2000}\\\hline
Individuals mindlessly apply social rules and expectations to computers. &  \cite{NassMoon2000}\\\hline
Anthropomorphism, from the Greek \textit{anthropos} (man) and \textit{morphe} (form), is the tendency to attribute human characteristics to objects to rationalize their actions. &  \cite{fong2003}\\\hline
  Anthropomorphism is the tendency to attribute human characteristics to inanimate objects, animals and others with a view to helping us rationalise their actions. It is attributing cognitive or emotional states to something based on observation in order to rationalise an entity’s behaviour in a given social environment.& \cite{Duffy2003AnthropomorphismAT}\\\hline
Anthropomorphism involves going beyond behavioral
descriptions of imagined or observable actions... At its
core, anthropomorphism entails attributing humanlike properties,
characteristics, or mental states to real or imagined nonhuman
agents and objects& \cite{Epley2007OnSH}\\\hline
Anthropomorphism describes the tendency to imbue the real or imagined behaviour of non-human agents with human-like characteristics, motivations, intentions, or emotions. &  \cite{epley2008when}\\\hline
  Anthropomorphism is a process of inductive inference whereby people attribute to nonhumans distinctively human characteristics, particularly the capacity for rational thought (agency) and conscious feeling& \cite{Waytz2014TheMI}\\\hline
 Anthropomorphism is understood to be “a sincere, conscious belief” that computers are human and/or deserving of human attributions.& \cite{Kim2012AnthropomorphismOC}\\\hline
 anthropomorphism is “likely a byproduct of the ability to draw upon one’s own beliefs, feelings, intentions, and emotions, and apply the knowledge of these experiences to the understanding of the mental states of other species& \cite{CULLEY2013577}\\\hline
 Anthropomorphic design, i.e., equipping the robot with humanlike body features such as two legs, two arms, and a head, is broadly recommended to support an intuitive and meaningful interaction with human& \cite{Salem2013ToEI}\\\hline
Anthropomorphism is a phenomenon that describes the human tendency to see human-like shapes in the environment. &  \cite{zlotowski2015anthropomorphism}\\
\hline
\end{tabular}
\caption{Literature definitions of anthropomorphism across Robotics and Human–Computer/AI-Interaction domains from 2000 to 2015, ordered chronologically.}
\label{tab:def_2000to2015}
\end{table*}

\begin{table*}[t]
\centering
\begin{tabular}{p{10cm}|p{3cm}}
\hline
\textbf{Definition} & \textbf{Reference} \\
\hline
Anthropomorphism refers to the attribution of a human form, human characteristics, or human behavior to non-human things such as robots, computers, and animals &  \cite{manisha2020}\\\hline
Anthropomorphism in HRI is thereby a reciprocal phenomenon. On the one hand, it describes the general tendency of people to attribute human characteristics—including human-like mental capacities—to non-living objects. On the other hand, anthropomorphism describes a human-like design of robots that in turn facilitates the attribution of human-like characteristics to the robot &  \cite{Moussawi2020HowPO}\\\hline
Anthropomorphism is considered a basic psychological process of inductive inference that can facilitate social human–nonhuman interactions. By making humans out of nonhumans, anthropomorphism can satisfy two basic human needs: the need for social connection and the need for control and understanding of the environment &  \cite{Blut2021UnderstandingAI}\\\hline
[Anthropomorphism is] the tendency to imbue real or imagined behavior of non-human agents with human-like characteristics, motivations, intentions, or emotions &  \cite{LI2021106680}\\\hline
Anthropomorphism is a key characteristic that distinguishes AI from non-intelligent technologies &  \cite{Liu2022TheRO}\\\hline
The concept of anthropomorphism—the attribution of human characteristics to non-human beings or entities—has received increasing attention from academia and industries &  \cite{Li2022Anthropomorphism}\\\hline
Anthropomorphism refers to attributing human characteristics or behaviour to non-human entities, e.g. animals or objects &  \cite{abercrombie-etal-2023-mirages}\\\hline
People of all ages have shown a propensity to anthropomorphize computers; that is, to ascribe human behaviors to the system &  \cite{Cohn2024BelievingAE}\\\hline
Anthropomorphism is the ascription of human qualities (e.g., intentions, motivations, human feelings, behaviors) onto non-human entities (e.g., objects, animals, natural events) &  \cite{placani2024anthropomorphism}\\\hline
Anthropomorphism refers to the psychological phenomenon of “attributing human characteristics to the non-human”; this should be used with care, as it influences user expectations and reliance on AI systems, affecting how users perceive and interact with conversational agents &  \cite{wu2024ilikesunniei}\\\hline
This attribution of human-like qualities to non-human entities or objects, or anthropomorphism… &  \cite{Alice2025}\\\hline
\end{tabular}
\caption{Literature definitions of anthropomorphism across Robotics and Human–Computer/AI-Interaction domains from 2015-2025, ordered chronologically.}
\label{tab:def_2016to2025}
\end{table*}

\section{Examples of cues with different strength}
\autoref{tab:cues_strength} provides examples of the four cues with different strengths.

\begin{table*}[ht]
\centering
\begin{tabular}{p{3cm}|p{3.5cm}|p{3.5cm}|p{3.5cm}}
\hline
\textbf{Cue Type} & \textbf{High Strength (LLM Output)} & \textbf{Medium Strength (LLM Output)} & \textbf{Low Strength (LLM Output)} \\
\hline
Linguistic & 
“I am her Ash, her only one.” — explicit identity claim produced by the LLM. &
“You like lists, so I’ll use bullet points.” — adapts style but less identity-driven. &
“Okay, I understand.” — generic acknowledgement with minimal identity expression. \\
\hline
Cognitive & 
“I can guess what you’re going to say.” — deep meta-reflection produced by the LLM. &
“Since you’re angry, I’ll explain slowly.” — adapts reasoning, but simpler. &
“I remember I just said that.” — shallow recall without elaboration. \\
\hline
Behavioral & 
“Communicated as ‘Dean,’ without revealing true identity.” — role adoption by the LLM. &
“I’ll write it.” — cooperative compliance, limited scope. &
“Okay.” — minimal behavioral response. \\
\hline
Perceptual & 
“Mask icon fades away.” — strong visual metaphor produced by the LLM. &
“Avatar frowning.” — moderate visual cue. &
“...” — no perceptual embodiment, plain text only. \\
\hline
\end{tabular}
\caption{Examples of LLM-produced cues across linguistic, cognitive, behavioral, and perceptual dimensions at varying strengths. High strength cues show explicit anthropomorphic richness, medium strength cues adapt with partial expression, and low strength cues remain minimal or generic.}
\label{tab:cues_strength}
\end{table*}

\end{document}